\documentclass[review]{elsarticle}
\usepackage{siunitx}
\usepackage{pgfplots}
\usepackage{lineno,hyperref}
\usepackage[]{algorithm2e}
\biboptions{sort&compress}

\modulolinenumbers[5]
\usepackage{fixmath}
\usepackage{mathtools}
\usepackage{amsmath}
\usepackage{tabularx}
\usepackage{tikz}
\usetikzlibrary[patterns]
\usetikzlibrary{arrows,positioning,shapes} 
\usetikzlibrary{intersections, pgfplots.fillbetween}
\usepackage{float}
\usepackage{mwe}
\usepackage{tikzscale}
\usepackage{subcaption} 
\usepackage{gensymb}
\usetikzlibrary{calc} 
\usetikzlibrary{shapes,arrows,positioning,shadows,trees}
\usetikzlibrary{decorations.pathreplacing}
\usetikzlibrary{positioning}
\usetikzlibrary{shapes.misc}
\usetikzlibrary{patterns}
\usetikzlibrary{3d}
\usetikzlibrary{spy}
\usetikzlibrary[patterns]
\usetikzlibrary{arrows,positioning,shapes} 
\usepackage[toc,page]{appendix}
\usepackage{amssymb}
\usepackage[export]{adjustbox}
\usepackage{graphicx}
\usepackage{xcolor,colortbl}

\usepackage{listings}
\pgfplotsset{grid style={solid,white!90!black}}

\journal{Journal of Computational Science}

\newcommand{\new}{\color{black}}

\makeatletter
\def\ps@pprintTitle{%
	\let\@oddhead\@empty
	\let\@evenhead\@empty
	\def\@oddfoot{\reset@font\hfil\thepage\hfil}
	\let\@evenfoot\@oddfoot
}
\makeatother

\begin{document}
	
	\begin{frontmatter}
		\title{A Deep Learning Approach to the Inversion of Borehole Resistivity Measurements}
		
		\author
		{M. Shahriari (1), D. Pardo (2, 1, 3), A. Pic\'on (4, 2), A. Galdr\'an (5) \\  J. Del Ser (4, 1, 2), C. Torres-Verd\'in (6)\\
			{\footnotesize (1) Basque Center for Applied Mathematics, (BCAM), Bilbao, Spain.\\
			(2) University of the Basque Country (UPV/EHU), Leioa, Spain.\\
			(3) Ikerbasque (Basque Foundation for Sciences), Bilbao, Spain.\\
			(4) Tecnalia, Bilbao, Spain.\\
			(5) INESC TEC, Porto, Portugal.\\
			(6) The University of Texas at Austin, USA.}
		}
		
		\begin{abstract}
		Borehole resistivity measurements are routinely employed to measure the electrical properties of rocks penetrated by a well and to quantify the hydrocarbon pore volume of a reservoir. Depending on the degree of geometrical complexity, inversion techniques are often used to estimate layer-by-layer electrical properties from measurements. When used for well geosteering purposes, it becomes essential to invert the measurements into layer-by-layer values of electrical resistivity in real time. We explore the possibility of using deep neural networks (DNNs) to perform rapid inversion of borehole resistivity measurements. Accordingly, we construct a DNN that approximates the following inverse problem: given a set of borehole resistivity measurements, the DNN is designed to deliver a physically reliable and data-consistent piecewise one-dimensional layered model of the surrounding subsurface. Once the DNN is constructed, we can invert borehole measurements in real time. We illustrate the performance of the DNN for inverting logging-while-drilling (LWD) measurements acquired in high-angle wells via synthetic examples. Numerical results are promising, although further work is needed to achieve the accuracy and reliability required by petrophysicists and drillers.
		\end{abstract}
		
		
	\end{frontmatter}
	
	\section{Introduction}
	One of the purposes of geophysical measurements is to interrogate the subsurface of the Earth to find oil and gas, and to optimize the production of existing hydrocarbon reservoirs. We divide existing geophysical measurements into two categories: (a) surface geophysical measurements, such as controlled source electromagnetics (CSEM) (see, e.g., \cite{Constable,Bakr}), seismic (see, e.g., \cite{Bob}), and magnetotellurics (MT) (see, e.g., \cite{Aramberri}), and (b) borehole sensing, such as logging-while-drilling (LWD) data (see, e.g., \cite{Davydycheva,Ijasan}). 
	
\begin{figure*}[ht]
		\centering
		\includegraphics{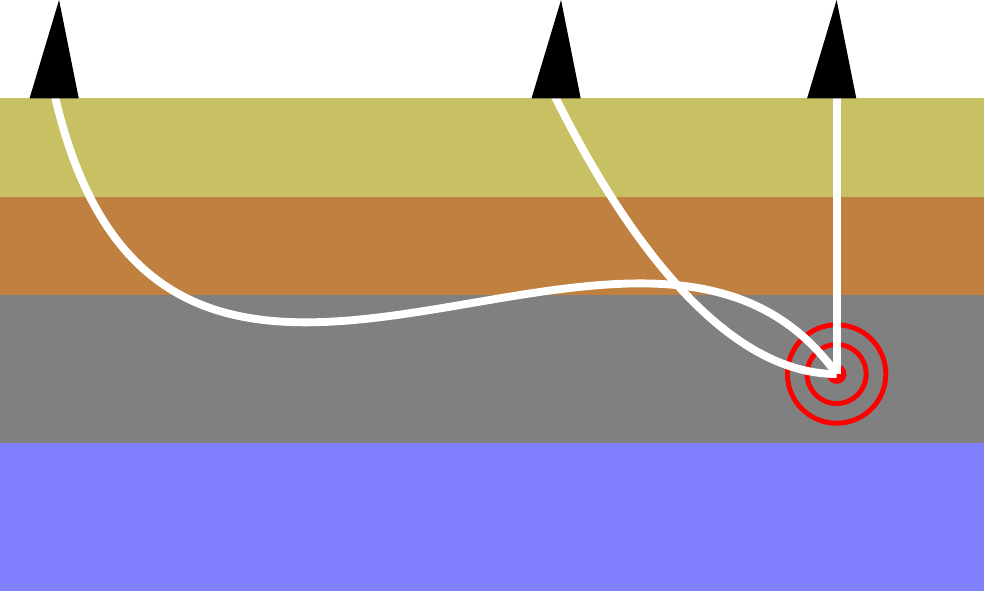}
		\caption{Drawing of three wells showing how different well trajectories can be employed to reach a specific subsurface target.}
		\label{fig_geo}
	\end{figure*}

	In this paper, we focus on borehole resistivity measurements. In particular, on those acquired with LWD instruments, which are currently widely used for well geosteering applications (see Figure \ref{fig_geo}). These logging instruments are equipped with one or various transmitters that emit electromagnetic waves, which are recorded at receivers that are also mounted on the same logging device. By adequately interpreting (inverting) these measurements, it is possible to determine the subsurface electromagnetic properties nearby the well, thus enabling the selection of an optimal well trajectory to target hydrocarbon-producing zones.
	
\begin{figure*}[ht]
		\centering
\begin{tikzpicture}

\draw[fill=white!70!blue] (-4.,3) -- (-0.5,3) -- (-0.5,5.6) -- (-4,5.6) -- (-4,3);
\draw[fill=black!30!white] (2.5,3.) -- (6,3.) -- (6,5.6) -- (2.5,5.6) -- (2.5,3.);
\node (z5) at (4.25,6.) {Output};
\node (z3) at (-2.25,6.) {Input};
\node (z1) at (-5.,4.3) {Forward:};
\node[align=center] (z7) at (-2.25,4.35) {Subsurface \\ properties ${\bf P}$ \\ + \\ Well trajectory ${\bf T}$}; 
\node[align=center] (z10) at (4.25,4.3) {Measurements $\boldsymbol{\cal M}$};
\draw [->,line width=2] (-0.4,4.3) -- (2.4,4.3); 
\node[align=center] (z11) at (1,4.) {{\bf F}};

\draw[fill=white!70!blue] (-4,-1.1) -- (-0.5,-1.1) -- (-0.5,1.5) -- (-4,1.5) -- (-4,-1.1);
\draw[fill=black!30!white] (2.5,-1.1) -- (6,-1.1) -- (6,1.5) -- (2.5,1.5) -- (2.5,-1.1);
\node (z2) at (-5.,0.2) {Inverse:};
\node (z4) at (-2.25,2.) {Input};
\node (z6) at (4.25,2.) {Output};
\node[align=center] (z8) at (-2.25,0.2) {Measurements $\boldsymbol{\cal M}$ \\ + \\ Well trajectory ${\bf T}$};
\node[align=center] (z9) at (4.25,0.2) {Subsurface \\ properties ${\bf P}$ };
\draw [->,line width=2] (-0.4,0.2) -- (2.4,0.2); 
\node[align=center] (z12) at (1,-0.1) {{\bf I}};
\end{tikzpicture} 
		\caption{High-level description of a forward and an inverse problem.}
		\label{pro_pic}
	\end{figure*}
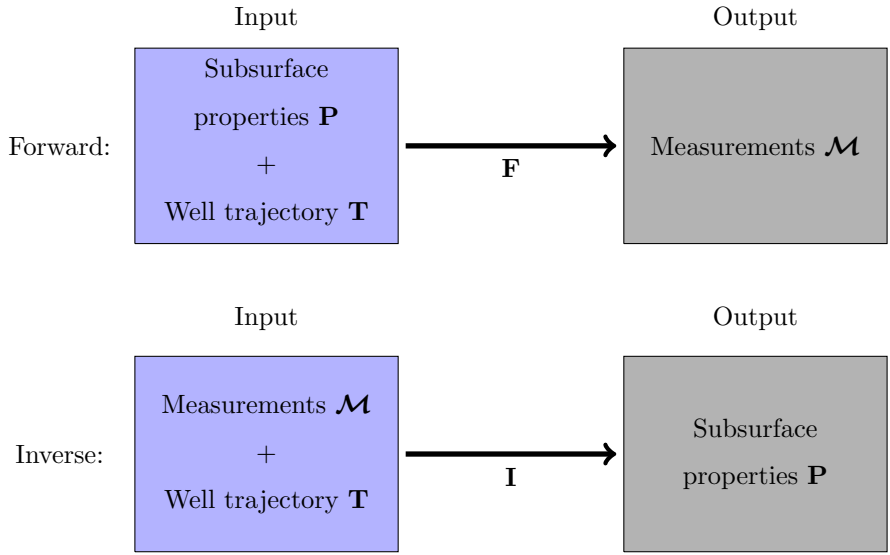

	From the mathematical point of view, we identify two different problems depicted in Figure \ref{pro_pic}:
	
	\begin{itemize}
		\item {\bf Forward problem}: Given a transmitter $t$ and known material properties ({\new in our case, a resistivity distribution and the geometrical characteristics of the media represented by vector ${\bf p}$}), the forward problem delivers the magnetic field (or a post-processed quantity of it) denoted by ${\bf m}$ (a vector of measurement) at a receiver $r$. Denoting by ${\bf T}$ a well trajectory composed of several logging positions (i.e., ${\bf T}=\{{\bf t}_i\}_{i=1}^{T}$, where $T$ is the number of logging positions), we have:
		\begin{equation}
		{\bf M} = \boldsymbol{\cal F} ({\bf p}; {\bf T}), 
		\end{equation}
		where $\boldsymbol{\cal F}$ accounts for a partial differential equation (PDE) based on Maxwell's equations and boundary conditions governing the electromagnetic wave propagation phenomena, and ${\bf M}=\{{\bf m}_i\}_{i=1}^M$ is the vector of measurements acquired along the well trajectory ${\bf T}$, where $M$ is the number of measurements (see e.g., \cite{Davydycheva,Davydycheva1,Shahriari}).
		
		\item {\bf Inverse problem}: Given a set of measurements ${\bf M}$ obtained over a specified logging trajectory ${\bf T}$, the solution of the inverse problem delivers a material subsurface distribution ${\bf p} \in \mathbb{R}^P $ (see, e.g., \cite{Ijasan,Pardo,Key}), where $P$ is the number of Parameters characterizing the media. An analytical expression of the governing equation $\boldsymbol{\cal I}$ that relates these variables is unknown. Nonetheless, for convenience, we express this problem as:
		\begin{equation}
		{\bf p}=\boldsymbol{\cal I} ({\bf M}; {\bf T}). 
		\end{equation}
		Mathematically speaking, the above function $\boldsymbol{\cal I}$ is not well-defined. For a given set of input parameters, it may have no output or, as it occurs more frequently, it can provide multiple outputs. These well-known undesirable properties of inverse problems (see, e.g., \cite{Tarantola,Vogel}) make them much more difficult to treat than forward problems. Various techniques such as regularization are intended to overcome these challenges and simplify the solution of inverse problems. The incorporation of non-linear constraints into $\boldsymbol{\cal I}$ is also a common technique to prevent non-physical solutions (see, e.g., \cite{Tarantola}).
	\end{itemize}
	
	An inversion problem is mathematically posed as a minimization of a cost function. There exist multiple approaches in the literature to solve such minimization problems. A popular one is based on the use of gradient-based algorithms \cite{Tarantola,Vogel}. However, they only guarantee a local minimum, which can be far away from the global one. Another family of methods is based on statistical algorithms \cite{Tarantola,Vogel,Watzenig}. However, they often require a large number of simulations, which increases the computational time. Moreover, for each new dataset of measurements, one needs to repeat the entire inversion process, which could be computationally intensive. This occurs because none of these methods deliver a full approximation of function $\boldsymbol{\cal I}$ itself, but rather they evaluate it over a particular set of measurements.
	
	In this work, we propose a different approach based on approximating function $\boldsymbol{\cal I}$ {\em offline} (i.e., {\em a priori}) using a deep neural network (DNN), and then, during field operations ({\em online}), evaluating this approximation for each given set of recorded measurements {${\bf M}$}.
	
	Early DNNs were already proposed in 1965 \cite{Ivakhnenko}. The term \textit{deep learning} was introduced in 1986 \cite{Dechter}, and later in 2000 \cite{Aizenberg} to refer to neural networks (NNs) that contain a large number of layers \cite{Lu}. A DNN enables to automatically detect and extract complex features that may be present in a given dataset. This was not possible with traditional NNs. In the last decade, DNNs have proven to be useful in multiple areas of knowledge (including computer vision \cite{Lu}, speech recognition \cite{Yu}, and biometrics \cite{Kumar}) to approximate complex functions with unknown properties. In recent years, the use of machine learning algorithms \cite{Bougher,Araya-Polo,Lary,Hedge,Aulia,Bize-Forest} and deep learning \cite{Wang,Higham} in computational mechanics and computational geophysics has become an active area of study. However, to the best of our knowledge, deep learning algorithms have not been applied to the inversion of borehole resistivity measurements, and therefore, its advantages and limitations on this area are unexplored. 
	
	In this work, we provide an introduction for geophysicists on the use of DNNs for solving inverse problems and analyze their main features and limitations when applied to the rapid interpretation of borehole resistivity measurements for geosteering purposes. {\new To simplify the problem and increase the speed of computations, we restrict to Earth formations composed by a sequence of one-dimensional (1D) layers, as described in \cite{Pardo}. The use of this assumption is common in the oil \& gas industry for the inversion of borehole resistivity measurements  \cite{Ijasan,Key1}.}
	
	The remaining part of this document is organized as follows. Section \ref{sec:deep_learning} provides an introduction to deep learning algorithms. Section \ref{sec:geverning_eq} describes the governing equation for borehole resistivity measurements. We introduce our measurement acquisition system in Section \ref{sec:measurements}. Section \ref{sec:traj_par} explains the parameterization (discretization) we select for the well trajectory. A similar description for the material properties discretization is provided in Section \ref{sec:mat_par}. Section \ref{sec:training} describes the training of our DNNs, and it shows the results of the training stage. Section \ref{sec:inversion_results} demonstrates the applicability of DNNs for inversion of borehole resistivity measurements via synthetic examples. Finally, Section \ref{sec:conclusion} is devoted to conclusions and future work. We also include three appendices that describe some advanced technical details about the DNN employed in this work.
	\section{Deep Neural Networks for Inverting Resistivity Measurements}
	\label{sec:deep_learning}
	In this section, we consider a discrete representation of the inverse function \\$\boldsymbol{\cal I}_h:\mathbb{R}^M \times \mathbb{R}^{3T} \rightarrow \mathbb{R}^P$ that associates each pair of measurements and trajectories $({\bf M} \times {\bf T}) \in \mathbb{R}^M \times \mathbb{R}^{3T}$ with a corresponding distribution of subsurface properties ${\bf p} \in \mathbb{R}^P$. In order to approximate this function, we employ NNs \cite{hornik_approximation_1991}. We provide below a concise overview of how to construct this kind of operators. The existing literature about NNs is large, but in here we only intend to briefly introduce some NNs and related algorithms to geophysicists that are relevant for the inversion of borehole resistivity measurements.
	
	\subsection{Fully-Connected Neural Network}\label{nn}
	Early formulations of NNs, known as fully-connected neural networks (FC-NNs), were defined by repeated compositions of simple transformations. Denoting ${\bf x}=({\bf M}, {\bf T})$, an FC-NN composed of $L$ \textit{layers} is given by:
	\begin{equation}\label{nn_eq}
	\boldsymbol{\cal I}_{\boldsymbol{\theta}}({\bf x}) = (\boldsymbol{\mathcal{N}}^{(L)} \circ \ldots \circ \boldsymbol{\mathcal{N}}^{(l)}\circ \ldots \boldsymbol{\mathcal{N}}^{(2)} \circ \boldsymbol{\mathcal{N}}^{(1)})({\bf x}),
	\end{equation}
	where $\boldsymbol{\mathcal{N}}^{(l)}({\bf x}) = \mathbf{s}({\bf W}^{(l)} \cdot {\bf x} + {\bf b}^{(l)})$ , ${\bf W}^{(l)}$ is a matrix, and ${\bf b}^{(l)}$ a vector. Thus, ${\bf W}^{(l)} \cdot {\bf x} + {\bf b}^{(l)}$ is an affine transformation. ${\bf s}$ is a simple non-linear point-wise mapping (activation function), typically the so-called \textit{rectified linear unit} given by:
	\begin{equation}\label{activation}
	\mathbf{s}({v_1,...,v_r})=(\max(0,v_1),...,\max(0,v_r)).
	\end{equation}
	Other activation functions can be applied, with arguably worse gradient preserving properties (e.g., tanh). We define $\boldsymbol{\theta}^{(l)}$ as a vector composed of all entries of matrix ${\bf W}^{(l)}$ and vector ${\bf b}^{(l)}$ for each layer $l=1,...,L$. Thus, $\boldsymbol{\theta} = \{\boldsymbol{\theta}^{(l)} \ : \ 1\leq l \leq L\}$ is a large vector of parameters fully determining $\boldsymbol{\cal I}_{\boldsymbol{\theta}}$. 
	Due to the varying dimensions of the different matrices ${\bf W}^{(l)}$ and vectors ${\bf b}^{(l)}$ at each layer in Equation \eqref{nn_eq}, the dimensionality of the input ${\bf x}$ can change, eventually reaching that of the target variable ${\bf p \in \mathbb{R}^P}$.
	

	\subsection{Training an NN: Data Preparation}
	We consider a finite set ${\cal S}$ containing $m$ data samples:
	\begin{equation}\label{global_set}
	\begin{split}
	&{\cal S}=({\cal M}, {\cal T}, {\cal P}) = \{({\bf M}^{[i]}, {\bf T}^{[i]}, {\bf p}^{[i]}) \ :\\
	& \  {\bf M}^{[i]} \in \mathbb{R}^M, {\bf T}^{[i]} \in \mathbb{R}^{3T}, {\bf p}^{[i]}\in \mathbb{R}^P, \ 1\leq i \leq m\}.
	\end{split}
	\end{equation}
	This set is randomly split into three disjoint subsets, referred to as \textit{training}, \textit{validation}, and \textit{test} sets, respectively:
	\begin{equation}
	\begin{split}
	{\cal S}_{train} &= ({\cal M}_{\mathrm{train}},{\cal T}_{\mathrm{train}},{\cal P}_{\mathrm{train}}) \\
	&= \{({\bf M}^{[i]}, {\bf T}^{[i]}, {\bf p}^{[i]}) \ : \ 1\leq i \leq m_1\},\\
	{\cal S}_{val} &= ({\cal M}_{\mathrm{val}},{\cal T}_{\mathrm{val}},{\cal P}_{\mathrm{val}})\\
	& = \{({\bf M}^{[i]}, {\bf T}^{[i]}, {\bf p}^{[i]}) \ : \ m_1+1\leq i \leq m_2\},\\
	{\cal S}_{test} &= ({\cal M}_{\mathrm{test}},{\cal T}_{\mathrm{test}},{\cal P}_{\mathrm{test}}) \\
	&= \{({\bf M}^{[i]}, {\bf T}^{[i]}, {\bf p}^{[i]}) \ : \ m_2+1\leq i \leq m\}.
	\end{split}
	\end{equation}
	We apply a network $\boldsymbol{\cal I}_{\boldsymbol{\theta}}$ to the input data sampled from set ${\cal S}$ in order to produce a prediction $\boldsymbol{\cal I}_{\boldsymbol{\theta}}({\bf M}^{[i]}, {\bf T}^{[i]})$ of its resistivity values. Then, one can compute the accuracy of such prediction via an error function $\mathcal{L}$, in our case given by the $\mathit{l}_2$ norm of the difference between both vectors:
	\begin{equation}\label{loss}
	\mathcal{L}(\boldsymbol{\cal I}_{\boldsymbol{\theta}}({\bf M}^{[i]}, {\bf T}^{[i]}), {\bf p}^{[i]}) = \|\boldsymbol{\cal I}_{\boldsymbol{\theta}}({\bf M}^{[i]}, {\bf T}^{[i]})-{\bf p}^{[i]}\|_{2}.
	\end{equation}
	The numerical process by which the error given by Equation \eqref{loss} is iteratively minimized via a gradient-based algorithm across the entire training set is referred to as \textit{training}, and will be detailed in the next subsection. The validation set ${\cal S}_{val}$ is employed to perform some high-level NN design decisions, e.g., to modify the network architecture (the dimensions of the different layers) or different parameters controlling the numerical optimization algorithm. After training, we compute predictions for data samples in ${\cal S}_{test}$ . Then, the network $\boldsymbol{\cal I}_{\boldsymbol{\theta}}$ is said to generalize properly if the errors in ${\cal S}_{train}$ and ${\cal S}_{test}$ are similar. In addition, if such errors are relatively low, we can assume that $\boldsymbol{\cal I}_{\boldsymbol{\theta}}$ correctly approximates operator $\boldsymbol{\cal I}_h$.

	\subsection{Training an NN: Numerical Optimization}
	A critical feature of NNs is that they are a hierarchical composition of multiple functions that are easy to differentiate. Hence, the chain rule becomes an essential tool to find derivatives of these operators. This is the core idea of the most popular algorithm for implementing gradient descent strategies on NNs, called \textit{back-propagation} in the NN's literature \cite{rumelhart_parallel_1986}.
	
	Within each gradient descent iteration, we first carry out a forward pass for a given data sample $(m_i, t_i)$ in order to compute a prediction $\boldsymbol{\cal I}_{\boldsymbol{\theta}}({\bf M}^{[i]}, {\bf T}^{[i]})$ and the corresponding error $\mathcal{L}(\boldsymbol{\cal I}_{\boldsymbol{\theta}}({\bf M}^{[i]}, {\bf T}^{[i]}),{\bf p}^{[i]})$. Afterwards, this error is backpropagated by applying the chain rule to the composition of functions defining the different layers of the network. Hence, proceeding from the last layer of the model backwards, one can estimate the gradient of the loss function with respect to parameters ${\boldsymbol{\theta}}$, defining $\boldsymbol{\cal I}_{\boldsymbol{\theta}}$ in reverse order as:
	\begin{equation} 
	\frac{\displaystyle \partial \mathcal{L}(\boldsymbol{\theta})}{\displaystyle \partial \boldsymbol{\theta}^{(L)}}, 
	\ldots, 
	\frac{\displaystyle \partial \mathcal{L}(\boldsymbol{\theta})}{\displaystyle \partial \boldsymbol{\theta}^{(l+1)}},
	\frac{\displaystyle \partial \mathcal{L}(\boldsymbol{\theta})}{\displaystyle \partial \boldsymbol{\theta}^{(l)}},
	\ldots,
	\frac{\displaystyle \partial \mathcal{L}(\boldsymbol{\theta})}{\displaystyle \partial\boldsymbol{\theta}^{(1)}}
	\end{equation}
	The gradient at each layer is derived based on previous gradient computations, parameters $\boldsymbol{\theta}$ are updated with some form of gradient descent strategy (e.g., stochastic gradient descent), and the process is iterated over all elements of the training set so as to minimize its average error. 
	
	The number of iterations during which the model is trained is typically decided by monitoring the value of the loss function $\mathcal{L}$ on elements of the validation set ${\cal S}_{val}$ that are never used to adjust the network parameters. 
	During training, that value is compared with the loss value attained in ${\cal S}_{train}$ in order to stop the optimization process as soon as both quantities start to diverge, which would imply that the network is becoming too much adjusted to the training data and failing to generalize, a phenomenon known as \textit{overfitting}.
	
	\subsection{Convolutional Neural Networks}
	As observed from Equation \eqref{nn_eq}, NNs are defined as a composition of functions. Thus, they naturally possess a layer-wise hierarchical nature. Therefore, they are ideal candidates to design operators that progressively retain the most salient aspects of the initial input. However, ${\bf W}^{(l)}$ are dense matrices, connecting every component of the input of a given layer to its output. This results in an excessively large number of parameters that need to be optimized. In order to reduce this number, a popular solution consists of replacing fully-connected affine layers $\boldsymbol{\mathcal{N}}$ by convolutional operators $\boldsymbol{\mathcal{C}}$ defined by convolution kernels $f$. This localizes computations, effectively reducing the number of parameters in $\boldsymbol{\cal I}_{\boldsymbol{\theta}}$. The resulting network is known as a convolutional neural network (CNN) \cite{lecun_gradient-based_1998}. We provide a rigorous definition of a CNN in Appendix \ref{convolutional}.
	
	\subsection{Recurrent Neural Networks}
	A particular kind of network architectures that are useful for sequence processing (e.g. speech, text, or time-related data) are recurrent neural networks (RNNs) \cite{hopfield_neural_1982}. In here, since successively recorded logging data exhibits a temporal pattern (there is a strong relationship between measurements recorded at a given logging position and at subsequent ones), we will also adopt an RNN design. For a technical description of this type of networks, see Appendix \ref{recurrent}.

	\subsection{An NN Architecture for Inverting Borehole Resistivity Measurements}
	The NN architecture employed in this work combines both a CNN and an RNN by first reducing the dimensionality of the input measurements employing a long short-term memory network, which is a specific class of RNN described in Appendix \ref{recurrent}. Next, the result of this operation serves as input to a series of one-dimensional CNNs, with interleaved \textit{pooling} operators similar to the ones described in Appendix \ref{convolutional}, where each convolutional block is based on a modified residual block \cite{He}, allowing deeper architectures while enhancing convergence. The output of this second set of operations becomes the input to a fully-connected layer that maps it into space $\mathbb{R}^P$ of subsurface resistivity properties. The network is trained end-to-end by backpropagation until the validation error is no longer decreasing. We provide a pseudo-code of this DNN in Appendix \ref{code}.
	\section{Reduced Wave Magnetic Field Equation}
	\label{sec:geverning_eq}
	Let ${\bf H}$ be the magnetic field, $\boldsymbol{{\cal M}}$ a magnetic source flux density, and $\boldsymbol{\sigma}=\boldsymbol \rho^{-1}$ a real-valued conductivity tensor with positive determinant. Then, the following reduced wave equation governs the magnetic field propagation phenomena:
	\begin{equation}
	\nabla \times \tilde{\boldsymbol{\sigma}}^{-1}\nabla \times {\bf H} - i\omega \boldsymbol{\mu} {\bf H}=i\omega \boldsymbol{\mu} \boldsymbol{{\cal M}}, 
	\label{Em}
	\end{equation}
	where $\tilde{\boldsymbol{\sigma}}^{-1}=(\boldsymbol{\sigma}-i\omega \boldsymbol{\varepsilon})^{-1}$, $\boldsymbol{\varepsilon}$ and $\boldsymbol{\mu}$ are the permittivity and magnetic permeability tensors of the media, respectively, $\omega=2 \pi f$ is the angular frequency, where $f>0$ is the frequency of operation of the transmitter, and $i$ is the imaginary unit, $i^2=-1$. The problem domain is $\Omega=\mathbb {R}^3$.
	
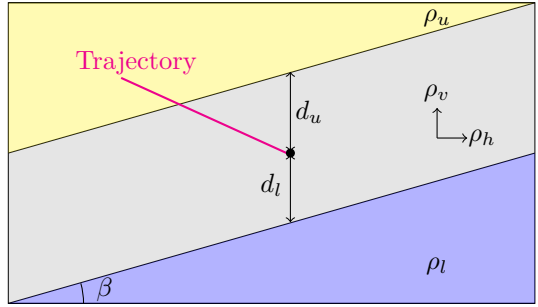
\begin{figure*}[ht]
		\centering
		\begin{tikzpicture}[scale=1.0]
		\node (layers) at (0,0)[scale=1]{
			\begin{tikzpicture}
\draw (0,0) rectangle (7,4);
\draw[name path=A1] (0,0) -- (7,0);
\draw[name path=A2] (0,4) -- (7,4);
\draw[name path=B] (0,2) -- (7,4);
\draw[name path=C] (0,0) -- (7,2);

\tikzfillbetween[of=A1 and C]{blue, opacity=0.3};
\tikzfillbetween[of= C and B]{gray, opacity=0.2};
\tikzfillbetween[of= B and A2]{yellow, opacity=0.3};

\draw[magenta, thick] (1.5,3) -- (3.7,2) node (n1) at (1.7,3.2) {Trajectory};
\fill[black] (3.75,2) circle (0.06cm);
\draw[<->] (3.75,2.02) -- (3.75,3.08);
\node (d_u) at (4,2.55) {$d_u$}; 
\draw[<->] (3.75,1.98) -- (3.75,1.08);
\node (d_u) at (3.5,1.58) {$d_l$}; 
\draw (1,0) arc (0:15.5:1 cm);
\node (beta) at (1.28,0.18) {$\beta$}; 
\node (rho_l) at (5.7,0.5) {$\rho_{l}$};
\draw [->] (5.7,2.2) -- (6.1,2.2);
\node (rho) at (6.3,2.2) {$\rho_h$};
\draw [->] (5.7,2.2) -- (5.7,2.6);
\node (rho) at (5.7,2.8) {$\rho_v$};
\node (rho_u) at (5.7,3.8) {$\rho_{u}$};
\end{tikzpicture} 
		};
		\end{tikzpicture}
		\caption{1D media and a trajectory. The black circle indicates the last position of the trajectory.}
		\label{1D}
	\end{figure*}

	In this work, we consider a sequence of 1D transversally isotropic (TI) media \cite{Pardo}. Therefore, the resistivity of the media varies only along $z$-direction (see Figure \ref{1D}), and we have: 
	\begin{equation}
	\boldsymbol \rho(z)=\left( 
	\begin{array}{ccc}
	\rho_h(z) & 0 & 0\\
	0 & \rho_h(z) & 0 \\
	0 & 0 & \rho_v(z)
	\end{array} \right),
	\end{equation}
	where $\rho_h$ and $\rho_v$ are the horizontal and vertical resistivities of the media, respectively.
	
	
	\section{Measurement Acquisition System} 
	\label{sec:measurements}   
\begin{figure*}[ht]
		\centering
		\begin{tikzpicture}[scale=1.0]
		\node (layers) at (0,0)[scale=1]{
			\begin{tikzpicture}

\fill[gray!80!white] (-1.5*3,0) -- (1.5*3,0.) --  node[right] {\small \bf \textcolor{black}{500 kHz}}  (1.5*3, 0.15*3) -- (-1.5*3,0.15*3) -- cycle;

\fill[black!80!white] (-0.6096*5-0.02*3,0) -- (-0.6096*5+0.02*3,0.) -- (-0.6096*5+0.02*3, 0.15*3) node[above] {\footnotesize \bf \textcolor{black}{Tx$_1$}}  -- (-0.6096*5-0.02*3,0.15*3) -- cycle;
\fill[black!80!white] (0.6096*5-0.02*3,0) -- (0.6096*5+0.02*3,0.)  -- (0.6096*5+0.02*3, 0.15*3) node[above] {\footnotesize \bf \textcolor{black}{Tx$_2$}} -- (0.6096*5-0.02*3,0.15*3) -- cycle;

\fill[red!80!white] (-0.1016*5-0.02*3,0) -- (-0.1016*5+0.02*3,0.)  -- (-0.1016*5+0.02*3, 0.15*3) node[above] {\footnotesize \bf \textcolor{black}{Rx$_1$}} -- (-0.1016*5-0.02*3,0.15*3) -- cycle;
\fill[red!80!white] ( 0.1016*5-0.02*3,0) -- ( 0.1016*5+0.02*3,0.) -- ( 0.1016*5+0.02*3, 0.15*3)  node[above] {\footnotesize \bf \textcolor{black}{Rx$_2$}} -- ( 0.1016*5-0.02*3,0.15*3) -- cycle;

\draw[black, line width=1pt,<->] (-0.1016*5, -0.6)      -- (0.1016*5, -0.6) node[pos=0.5, below] {\footnotesize \bf \textcolor{black}{0.40 m}}  ;
\draw[gray, dashed] (-0.1016*5, -0.6)      -- (-0.1016*5, 0)  ;
\draw[gray, dashed] ( 0.1016*5, -0.6)      -- ( 0.1016*5, 0)  ;

\draw[black, line width=1pt,<->] (-0.6096*5, -1.2)      -- (0.6096*5, -1.2) node[pos=0.5, below] {\footnotesize \bf \textcolor{black}{1.8 m}}  ;
\draw[gray, dashed] (-0.6096*5,-1.2)      -- (-0.6096*5, 0)  ;
\draw[gray, dashed] ( 0.6096*5,-1.2)      -- ( 0.6096*5, 0)  ;

\end{tikzpicture} 
		};
		\end{tikzpicture}
		\caption{Conventional LWD logging instrument. Tx$_i$ and Rx$_i$ are the transmitters and the receivers, respectively.}
		\label{fig_device_lwd}
	\end{figure*}
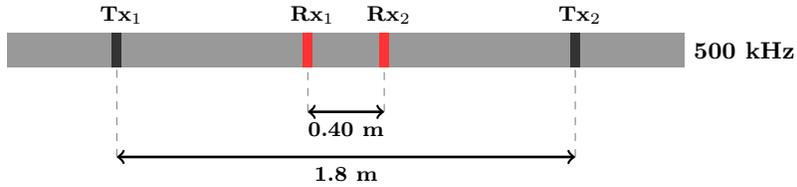

	In this work, we first consider the short co-axial LWD instrument as shown in Figure \ref{fig_device_lwd}. For this instrument, we measure attenuation and phase difference. We will denote those measurements as ${\cal M}_1$. To compute them, we consider the $zz$ coupling ${H}_{zz}$, where the first and the second subscripts correspond to the direction of the transmitter and the receiver, respectively. We record these quantities at both receivers, and denote them as ${H}^1_{zz}$ and ${H}^2_{zz}$. We define the attenuation and the phase difference as follows:
	\begin{equation}
	\begin{split}
	\ln \frac{{H}^1_{zz}}{{H}^2_{zz}}&= \underbrace{\ln \frac{\mid {H}^1_{zz}\mid}{\mid{H}^2_{zz}\mid}}_{\times 20 \log(e) = \text{attenuation }(dB)}\\
	&+i \underbrace{\left( ph({H}^1_{zz})-ph({H}^2_{zz})\right)}_{\times \dfrac{180}{\pi} =\text{phase difference (degree)}},
	\end{split}
	\label{atten_phase}
	\end{equation}
	where $ph$ denotes the phase of a complex number.
	
\begin{figure*}[ht]
		\centering
		\begin{tikzpicture}[scale=1.0]
		\node (layers) at (0,0)[scale=1]{
			\begin{tikzpicture}

\fill[gray!80!white] (-1.7*3,0) -- (1.7*3,0.) --  node[right] {\small \bf \textcolor{black}{10 kHz}}  (1.7*3, 0.15*3) -- (-1.7*3,0.15*3) -- cycle;

\fill[black!80!white] (-0.6096*5-0.02*3,0) -- (-0.6096*5+0.02*3,0.) -- (-0.6096*5+0.02*3, 0.15*3) node[above] {\footnotesize \bf \textcolor{black}{Tx}}  -- (-0.6096*5-0.02*3,0.15*3) -- cycle;
\fill[red!80!white] (0.6096*5-0.02*3,0) -- (0.6096*5+0.02*3,0.)  -- (0.6096*5+0.02*3, 0.15*3) node[above] {\footnotesize \bf \textcolor{black}{Rx}} -- (0.6096*5-0.02*3,0.15*3) -- cycle;



\draw[black, line width=1pt,<->] (-0.6096*5, -0.7)      -- (0.6096*5, -0.7) node[pos=0.5, below] {\footnotesize \bf \textcolor{black}{12 m}}  ;
\draw[gray, dashed] (-0.6096*5,-0.7)      -- (-0.6096*5, 0)  ;
\draw[gray, dashed] ( 0.6096*5,-0.7)      -- ( 0.6096*5, 0)  ;

\end{tikzpicture} 
		};
		\end{tikzpicture}
		\caption{Short-spacing deep azimuthal instrument. Tx and Rx are the transmitter and the receiver, respectively.}
		\label{fig_device_azim}
	\end{figure*}
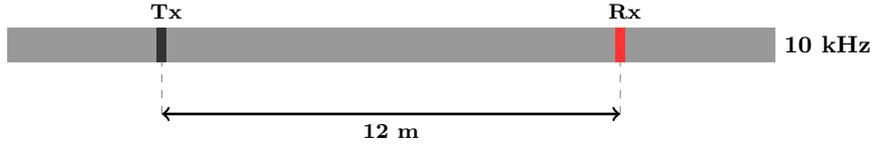

	In addition, we consider the short-spacing deep azimuthal instrument described in Figure \ref{fig_device_azim}. For this logging instrument, we record the attenuation and phase difference and denote these measurements as ${\cal M}_2$. We define them as in Equation \eqref{atten_phase} with ${H}^2_{zz}=1$ since there is no second transmitter. Finally, we also record a directional measurement referred as geosignal and defined as follows:  
	\begin{equation}
	\begin{split}
	g=&\ln \frac{{H}_{zz}-{H}_{zx}}{{H}_{zz}+{H}_{zx}}= \underbrace{\ln \frac{\mid {H}_{zz}-{H}_{zx}\mid}{\mid{H}_{zz}+{H}_{zx}\mid}}_{\times 20 \log(e) = \text{attenuation }(dB)}\\
	&+i \underbrace{\left( ph({H}_{zz}-{H}_{zx})-ph({H}_{zz}+{H}_{zx})\right)}_{\times \dfrac{180}{\pi} =\text{phase difference (degree)}}.
	\end{split}
	\end{equation}
	${\cal M}_3$ denotes the set of geosignal measurements.
	
	\section{Trajectory Parameterization}
	\label{sec:traj_par}
	We select a fixed number of tool positions based on the depth of investigation of the logging instruments. For our instruments, the largest depth of investigation is close to 20 $m$. By considering the logging step size equal to one foot (0.3048 m), we select $T=65$.
	
	We consider an arbitrary (but close to horizontal) trajectory, as it is customary in geosteering applications. Since we assume a 1D layered media on the proximity of the well trajectory, we select the azimuthal degree of the trajectory to be always equal to zero.
	
	With the above assumptions, we discretize (parameterize) the well trajectory as follows. We consider $\alpha_{ini}({\bf t})$ to be the initial trajectory dip angle. We assume that the trajectory dip angle can vary while drilling by an angle $\alpha_v$ in each step. Hence, at each tool position ($i$), the trajectory dip angle is:
	\begin{equation}
	\alpha({\bf t}_i)=\alpha_{ini}({\bf t})+(i-1) \alpha_v,\ i=1,\cdots, T,
	\end{equation}
	where $\alpha({\bf t}_i)$ is the trajectory dip angle at the $i$-th position.
	\section{Material Properties Parameterization}
	\label{sec:mat_par}
	For the successful 1D inversion of borehole resistivity measurements, it is often sufficient to recover a media containing only three layers for each logging position and it is characterized by the following seven variables: (A) the horizontal and vertical resistivity of the layer where the tool is currently located ($\rho_{h}$ and $\rho_{v}$, respectively); (B) the resistivity of the upper and lower layers located above and below the current logging position ($\rho_{u}$ and $\rho_{l}$, respectively); (C) the vertical distance from the current logging position to the upper and lower bed boundary positions ($d_{u}$ and $d_{l}$, respectively); and (D) the dip angle of the formation ($\beta$), which is assumed to be identical for all layers (see Figure \ref{1D}). Our DNN will provide an estimate of these seven numbers at each logging position.
	\section{Traning the DNN}    
	\label{sec:training}
	To produce reliable training and validation sets, and to avoid full randomness which may lead to non-physical data, we consider some physical and geological properties of the subsurface.
	\subsection{Material properties}
	In order to produce our training and validation sets, we select $\rho_u ,\rho_{l} \in [1,10^3]$. Since we want resistivity values to be comparable, we consider them in logarithmic scale. Thus, our random variables become: $\log(\rho_{l}),\log(\rho_{u}) \in [0,3]$. Additionally, in the case of $\rho_{h}$ and $\rho_{v}$, we incorporate the following physical restrictions:
	\begin{equation}
	\rho_{h} \le \rho_{v} \le 10 \rho_{h}.
	\label{cond}
	\end{equation}
	Therefore, we obtain:
	\begin{equation}
	1 \le \underbrace{\frac{\rho_{v}}{\rho_{h}}}_{a} \le 10 ,
	\end{equation}    
	where $a$ is the anisotropy factor. In order to impose restriction \eqref{cond} in our calculations, we select random values of $\log(a) \in [0,1]$. Moreover, since we want to have $\log(\rho_v) \in [0,3]$, we select $\log(\rho_h) \in [0,3-\log(a)]$. 
	We consider $d_{l},d_{u} \in [0.01,10]$ meters. As with the resistivity values, we consider them in logarithmic scale, i.e., our variables become $\log(d_{l}), \log(d_{u}) \in [-2,1]$. In geological layers, we assume that the dip angle is $\beta \in [-10 \degree,10 \degree]$.
	
	To summarize, we select $\log(\rho_u)$, $\log(\rho_l)$, $\log(\rho_h)$, $a$, $\log(d_{u})$, $\log(d_{l})$, and $\beta$ randomly within their aforementioned ranges of variation to characterize our synthetic forward models.
	\subsection{Trajectory}
	We consider an almost horizontal trajectory, as it occurs in most geosteering applications. Specifically, we restrict to $\alpha_{ini}({\bf t}) \in [83\degree,97\degree]$. Moreover, we further assume that the tool trajectory deviates by a maximum of $3\degree$ in a 20 meters section. In addition, since the direction of the trajectory dip angle is often changing gradually and almost constantly from one logging position to the next, for $T=65$ we have $\alpha_v \in [-0.045 \degree, 0.045\degree]$.
	
	By selecting randomly $\alpha_{ini}({\bf t})$ and $\alpha_v$ in their above ranges of variation, we build the trajectories for our forward problems.
	\begin{figure*}[ht]
		\centering
		\begin{tikzpicture}
		\node at (0,24.0)[scale=0.85]{\input{Model_2/Results/65/rho_h.tex}};
		\node at (4.7,24.0)[scale=0.85]{\input{Model_2/Results/65/a.tex}};
		\node at (9.4,24.0)[scale=0.85]{\input{Model_2/Results/65/rho_v.tex}};

		\node at (0,19.0)[scale=0.85]{\input{Model_2/Results/65/rho_u.tex}};
		\node at (4.7,19.0)[scale=0.85]{\input{Model_2/Results/65/rho_l.tex}};
		\node at (9.4,19.0)[scale=0.85]{\input{Model_2/Results/65/d_u.tex}};
		
		\node at (2.35,14.0)[scale=0.85]{\input{Model_2/Results/65/d_l.tex}};
		\node at (7.05,14.0)[scale=0.85]{\input{Model_2/Results/65/beta.tex}};
		\end{tikzpicture}
		\caption{Model problem 2. Comparison between the ground truth and predicted values using a trained DNN for ${\cal M}={\cal M}_2$. Red line indicates the equality of the predicted values and the ground truth. The blue lines correspond to the 10 and 90 percentiles, respectively.}
		\label{fig_results_m1}
	\end{figure*}
	\begin{figure*}[ht]
		\centering
		\begin{tikzpicture}
		\node at (0,24.0)[scale=0.85]{\input{Model_2/Results/130/rho_h.tex}};
		\node at (4.7,24.0)[scale=0.85]{\input{Model_2/Results/130/a.tex}};
		\node at (9.4,24.0)[scale=0.85]{\input{Model_2/Results/130/rho_v.tex}};

		\node at (0,19.0)[scale=0.85]{\input{Model_2/Results/130/rho_u.tex}};
		\node at (4.7,19.0)[scale=0.85]{\input{Model_2/Results/130/rho_l.tex}};
		\node at (9.4,19.0)[scale=0.85]{\input{Model_2/Results/130/d_u.tex}};
		
		\node at (2.35,14.0)[scale=0.85]{\input{Model_2/Results/130/d_l.tex}};
		\node at (7.05,14.0)[scale=0.85]{\input{Model_2/Results/130/beta.tex}};
		\end{tikzpicture}
		\caption{Model problem 2. Comparison between the ground truth and predicted values using a trained DNN using ${\cal M}={\cal M}_2 \cup {\cal M}_3$. Red line indicates the equality of the predicted values and the ground truth. The blue lines correspond to the 10 and 90 percentiles, respectively.}
		\label{fig_results_m2}
	\end{figure*}
	\begin{figure*}[ht]
		\centering
		\begin{tikzpicture}
		\node at (0,24.0)[scale=0.85]{\input{Model_2/Results/195/rho_h.tex}};
		\node at (4.7,24.0)[scale=0.85]{\input{Model_2/Results/195/a.tex}};
		\node at (9.4,24.0)[scale=0.85]{\input{Model_2/Results/195/rho_v.tex}};

		\node at (0,19.0)[scale=0.85]{\input{Model_2/Results/195/rho_u.tex}};
		\node at (4.7,19.0)[scale=0.85]{\input{Model_2/Results/195/rho_l.tex}};
		\node at (9.4,19.0)[scale=0.85]{\input{Model_2/Results/195/d_u.tex}};
		
		\node at (2.35,14.0)[scale=0.85]{\input{Model_2/Results/195/d_l.tex}};
		\node at (7.05,14.0)[scale=0.85]{\input{Model_2/Results/195/beta.tex}};
		\end{tikzpicture}
		\caption{Model problem 2. Comparison between the ground truth and predicted values using a trained DNN using ${\cal M}={\cal M}_1 \cup {\cal M}_2 \cup {\cal M}_3$. Red line indicates the equality of the predicted values and the ground truth. The blue lines correspond to the 10 and 90 percentiles, respectively.}
		\label{fig_results_m3}
	\end{figure*}
	\subsection{Results}
	{\new
		For experimental purposes, we generate one million randomly selected samples/trajectories and their associated formation models (80\% training, 10\% validation, and 10\% test). Figure \ref{fig_results_m1} shows the accuracy of the trained DNN when we only consider the set of measurements ${\cal M}_2$, i.e., ${\cal M}={\cal M}_2$. The red line indicates the perfect approximation where the predicted value and the ground truth (the real parameters associated with a formation) coincide. The upper and lower blue lines show percentiles 10 and 90, respectively. These percentiles provide a reliable uncertainty quantification. In a perfect approximation, the blue lines should coincide with the red one. Therefore, a lower distance between the blue lines and the red one indicates a better approximation. Figures show a denser cloud of points in the proximity of the red line, which indicates an acceptable approximation. However, for the anisotropy factor $a$, the DNN is almost unable to predict the correct value, and consequently, it can not predict $\rho_v$ as precisely as the other variables.
		
		Analogously, Figure \ref{fig_results_m2} illustrates the results when we select ${\cal M}={\cal M}_2 \cup {\cal M}_3$. One can see that the blue lines are closer compared to those shown in Figure \ref{fig_results_m1}. Moreover, the concentration of points in the proximity of the red line increases. However, the approximation of the anisotropy factor $a$ is still poor, although better than in the previous case.
		
		Figure \ref{fig_results_m3} illustrates the results when we employ all available measurements, i.e., ${\cal M}={\cal M}_1 \cup {\cal M}_2 \cup {\cal M}_3$. These results outperform previous ones obtained with fewer measurements and, for the first time, we obtain an acceptable prediction of anisotropy factor $a$.

		\section{Inversion Results}
		\label{sec:inversion_results}
		Since the DNN trained with measurements drawn from ${\cal M}={\cal M}_1 \cup {\cal M}_2 \cup {\cal M}_3$ exhibits the best performance, we use it to invert several practical synthetic examples.
		
		\begin{figure*}[ht]
			\centering
			\subcaptionbox{Actual formation}{%
				\includegraphics[scale=0.3]{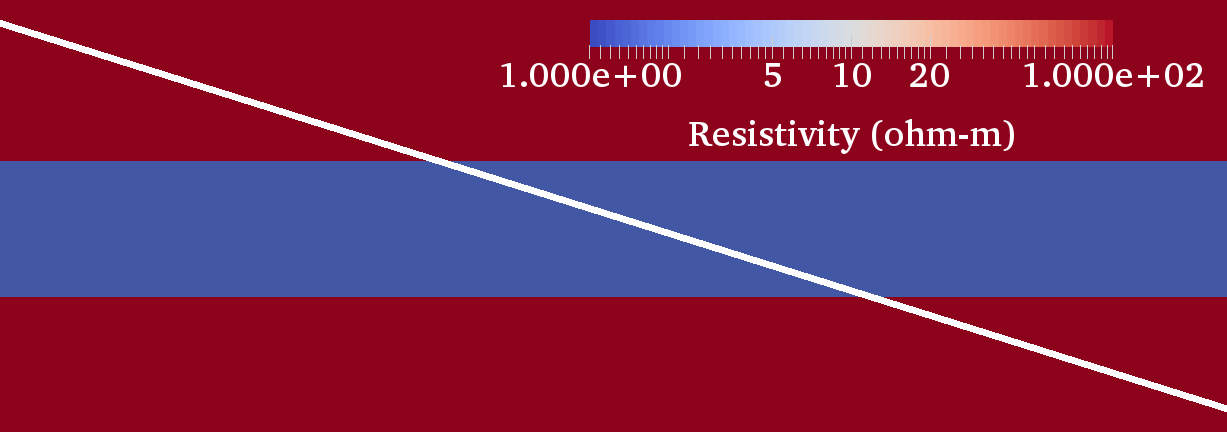} %
			}

			\bigskip
			\subcaptionbox{Predicted (inverted) formation}{%
				\includegraphics[scale=0.3]{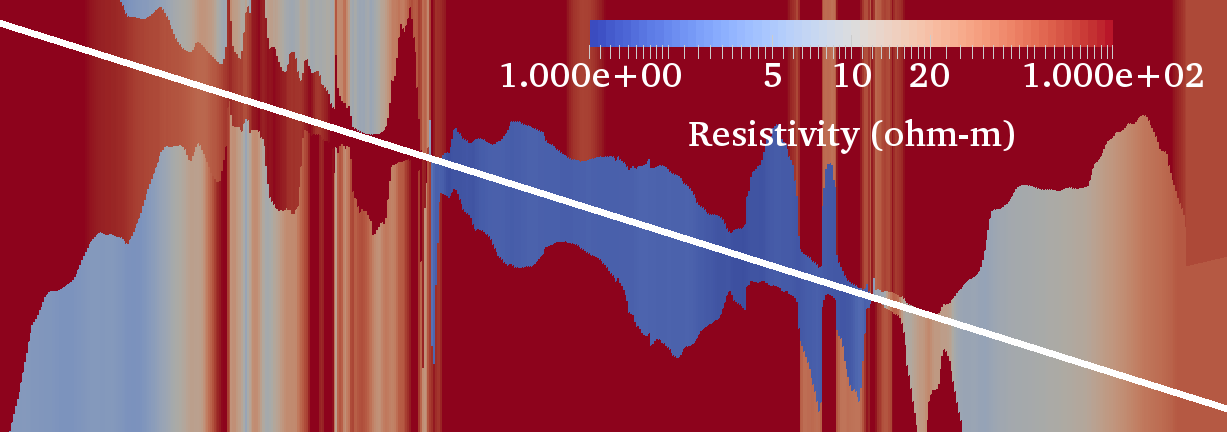}    
			}
			\caption{Model problem 1. Comparison between actual and predicted (inverted) formation.}
			\label{fig_M1}
		\end{figure*}
		\begin{figure*}[ht]
			\centering
			\subcaptionbox{${\cal M}_1$}{%
					\begin{tikzpicture}
	\begin{axis}[
	width=10cm,
	height=3 cm,
	scale only axis,
	enlargelimits=false,
	x label style={at={(axis description cs:0.5,0)},anchor=north},
	y label style={at={(axis description cs:0,0.5)},anchor=south},
	xlabel={True horizontal length (m)},
	ylabel={Attenuation (dB)},
	] 
	\addplot graphics[xmin=0,xmax=240,ymin=1.15,ymax=1.5] {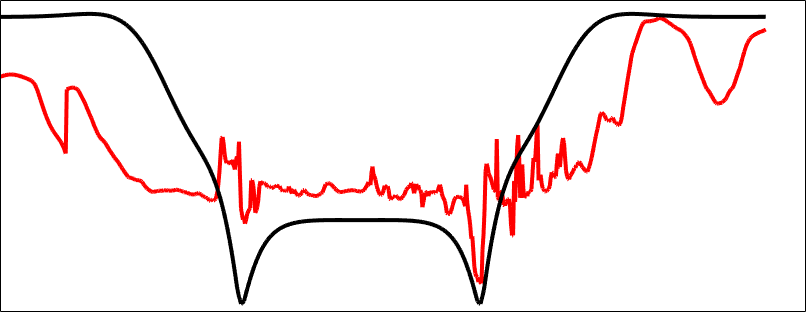};
	\end{axis}
    \node[align=center] (1) at (4.2,2)  { {\color{black} Exact}\\ {\color{red} Prediction}};
	\end{tikzpicture} %
			}

			\bigskip
			\subcaptionbox{${\cal M}_2$}{%
					\begin{tikzpicture}
	\begin{axis}[
	width=10cm,
	height=3 cm,
	scale only axis,
	enlargelimits=false,
	x label style={at={(axis description cs:0.5,0)},anchor=north},
	y label style={at={(axis description cs:0,0.5)},anchor=south},
	xlabel={True horizontal length (m)},
	ylabel={Attenuation (dB)},
	] 
	\addplot graphics[xmin=0,xmax=240,ymin=-7.7,ymax=-7.35] {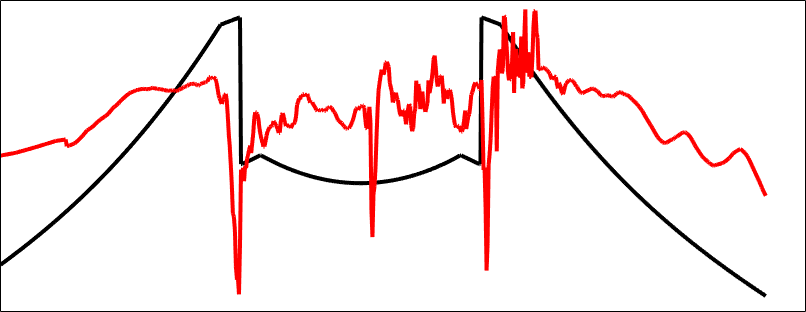};
	\end{axis}
    \node[align=center] (1) at (7,0.8)  { {\color{black} Exact}\\ {\color{red} Prediction}};
	\end{tikzpicture} %
			}

			\bigskip
			\subcaptionbox{${\cal M}_3$}{%
					\begin{tikzpicture}
	\begin{axis}[
	width=10cm,
	height=3 cm,
	scale only axis,
	enlargelimits=false,
	x label style={at={(axis description cs:0.5,0)},anchor=north},
	y label style={at={(axis description cs:0,0.5)},anchor=south},
	xlabel={True horizontal length (m)},
	ylabel={Attenuation (dB)},
	] 
	\addplot graphics[xmin=0,xmax=240,ymin=-0.8,ymax=1] {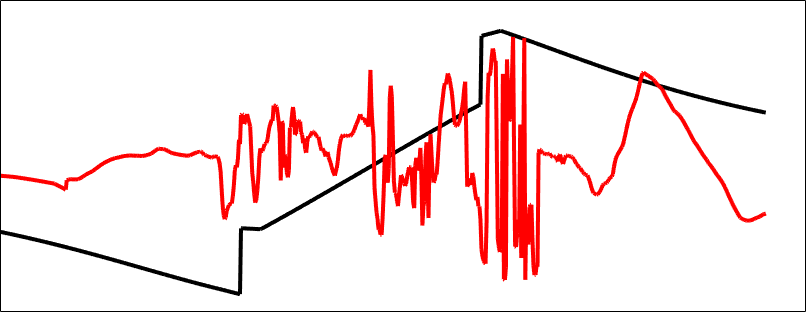};
	\end{axis}
    \node[align=center] (1) at (8.3,0.8)  { {\color{black} Exact}\\ {\color{red} Prediction}};
	\end{tikzpicture} %
			}
			\caption{Model problem 1. Comparison between exact and prediction of attenuation of the measurements.}
			\label{fig_M1_atten}
		\end{figure*}
		\begin{figure*}[ht]
			\centering
			\subcaptionbox{${\cal M}_1$}{%
				\begin{tikzpicture}
	\begin{axis}[
	width=10cm,
	height=3 cm,
		scale only axis,
		enlargelimits=false,
		x label style={at={(axis description cs:0.5,0)},anchor=north},
		y label style={at={(axis description cs:0,0.5)},anchor=south},
		xlabel={True horizontal length (m)},
		ylabel={Phase difference (degree)},
		] 
		\addplot graphics[xmin=0,xmax=240,ymin=-0.2,ymax=0.7] {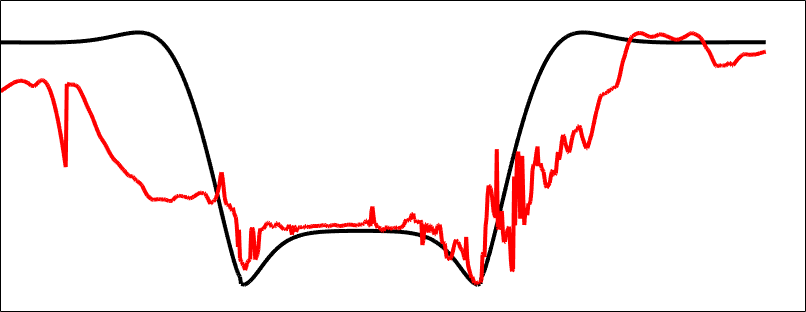};
	\end{axis}
    \node[align=center] (1) at (4.4,2)  { {\color{black} Exact}\\ {\color{red} Prediction}};
\end{tikzpicture}    
			}

			\bigskip
			\subcaptionbox{${\cal M}_2$}{%
				\begin{tikzpicture}
	\begin{axis}[
	width=10cm,
	height=3 cm,
		scale only axis,
		enlargelimits=false,
		x label style={at={(axis description cs:0.5,0)},anchor=north},
		y label style={at={(axis description cs:0,0.5)},anchor=south},
		xlabel={True horizontal length (m)},
		ylabel={Phase difference (degree)},
		] 
		\addplot graphics[xmin=0,xmax=240,ymin=-0.3,ymax=0.8] {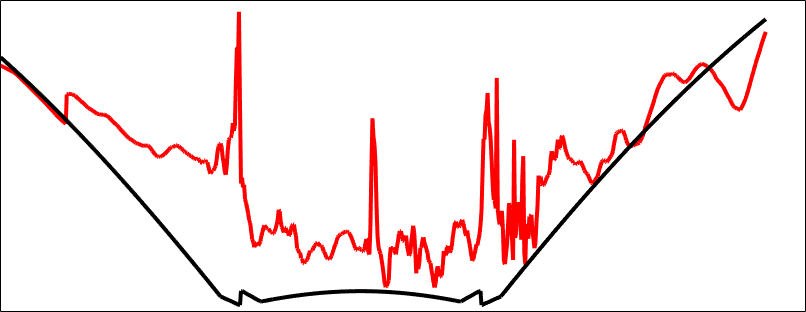};
	\end{axis}
    \node[align=center] (1) at (3.9,2.4)  { {\color{black} Exact}\\ {\color{red} Prediction}};
\end{tikzpicture}    
			}

			\bigskip
			\subcaptionbox{${\cal M}_3$}{%
				\begin{tikzpicture}
	\begin{axis}[
	width=10cm,
	height=3 cm,
		scale only axis,
		enlargelimits=false,
		x label style={at={(axis description cs:0.5,0)},anchor=north},
		y label style={at={(axis description cs:0,0.5)},anchor=south},
		xlabel={True horizontal length (m)},
		ylabel={Phase difference (degree)},
		] 
		\addplot graphics[xmin=0,xmax=240,ymin=-0.4,ymax=0.5] {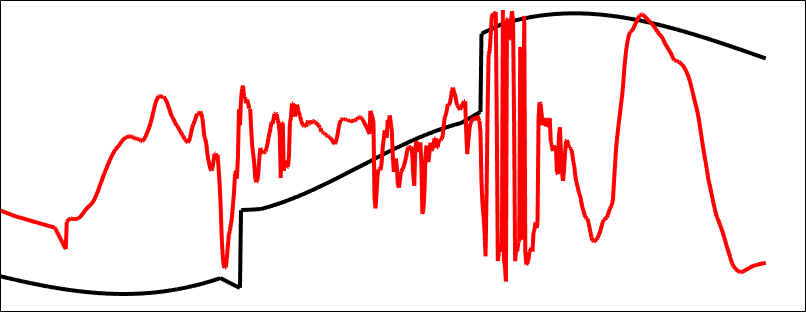};
	\end{axis}
    \node[align=center] (1) at (.9,2.4)  { {\color{black} Exact}\\ {\color{red} Prediction}};
\end{tikzpicture}    
			}
			\caption{Model problem 1. Comparison between exact and prediction of phase difference of the measurements.}
			\label{fig_M1_phase}
		\end{figure*}
		
		Figure \ref{fig_M1} illustrates the inversion of a three-layer media in which the middle layer is more conductive than the other ones, and it is anisotropic. Inversion results are less accurate than those possibly obtained with a gradient-based method. However, as initial results, they are encouraging. The results show that for the isotropic layer, the prediction of the resistivity is better than the one for the anisotropic layer. This probably occurs because the inversion of anisotropic factor $a$ presents some accuracy deficiencies. The predictions of $d_u$ and $d_l$ provide an acceptable view of the material surrounding the instrument. Figures \ref{fig_M1_atten} and \ref{fig_M1_phase} show a comparison between the attenuations and the phase differences of the measurements corresponding to the exact and predicted (inverted) models. These results show a better approximation of ${\cal M}_1$ than of ${\cal M}_2$ and ${\cal M}_3$.
		
		\begin{figure*}[ht]
			\centering
			\subcaptionbox{Actual formation}{%
				\includegraphics[scale=0.3]{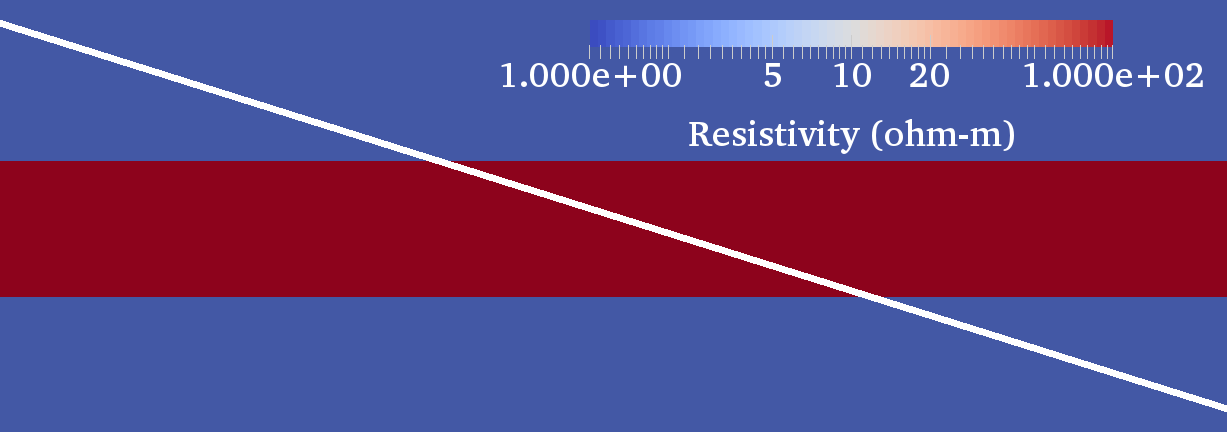} %
			}

			\bigskip
			\subcaptionbox{Predicted (inverted) formation}{%
				\includegraphics[scale=0.3]{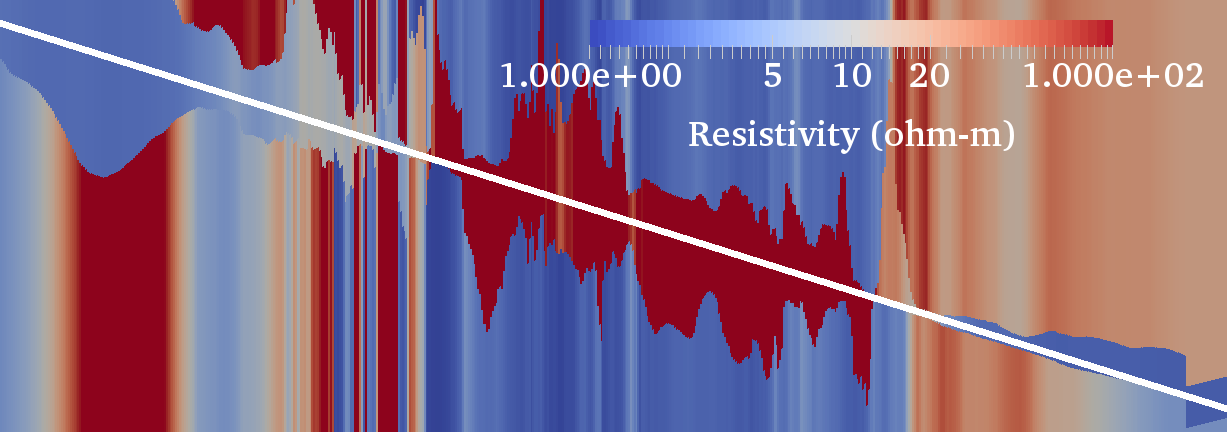}    
			}
			\caption{Model problem 2. Comparison between actual and predicted (inverted) formation.}
			\label{fig_M2}
		\end{figure*}
		\begin{figure*}[ht]
			\centering
			\subcaptionbox{${\cal M}_1$}{%
					\begin{tikzpicture}
	\begin{axis}[
	width=10cm,
	height=3 cm,
	scale only axis,
	enlargelimits=false,
	x label style={at={(axis description cs:0.5,0)},anchor=north},
	y label style={at={(axis description cs:0,0.5)},anchor=south},
	xlabel={True horizontal length (m)},
	ylabel={Attenuation (dB)},
	] 
	\addplot graphics[xmin=0,xmax=240,ymin=1.15,ymax=1.6] {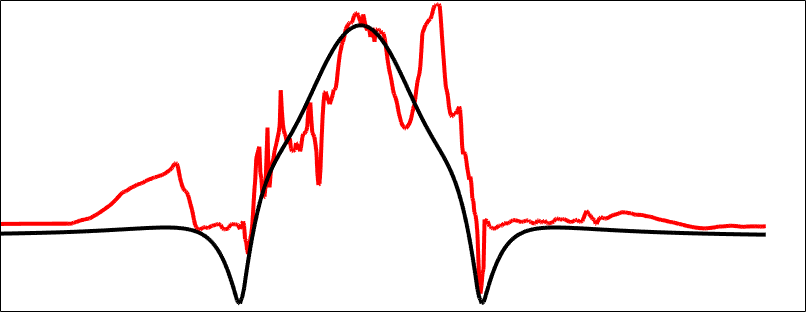};
	\end{axis}
    \node[align=center] (1) at (4.5,1.)  { {\color{black} Exact}\\ {\color{red} Prediction}};
	\end{tikzpicture} %
			}

			\bigskip
			\subcaptionbox{${\cal M}_2$}{%
					\begin{tikzpicture}
	\begin{axis}[
	width=10cm,
	height=3 cm,
	scale only axis,
	enlargelimits=false,
	x label style={at={(axis description cs:0.5,0)},anchor=north},
	y label style={at={(axis description cs:0,0.5)},anchor=south},
	xlabel={True horizontal length (m)},
	ylabel={Attenuation (dB)},
	] 
	\addplot graphics[xmin=0,xmax=240,ymin=-7.48,ymax=-7.34] {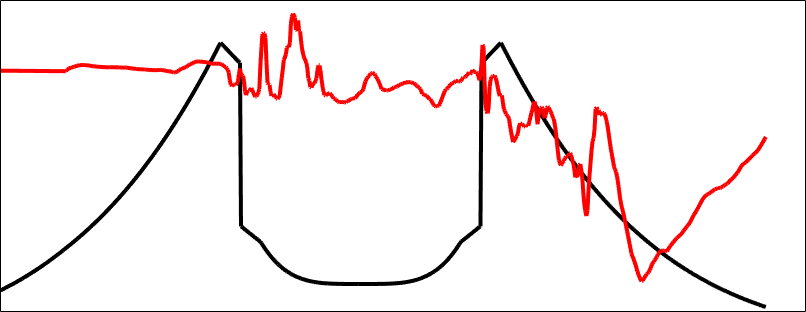};
	\end{axis}
    \node[align=center] (1) at (4.5,1.5)  { {\color{black} Exact}\\ {\color{red} Prediction}};
	\end{tikzpicture} %
			}

			\bigskip
			\subcaptionbox{${\cal M}_3$}{%
					\begin{tikzpicture}
	\begin{axis}[
	width=10cm,
	height=3 cm,
	scale only axis,
	enlargelimits=false,
	x label style={at={(axis description cs:0.5,0)},anchor=north},
	y label style={at={(axis description cs:0,0.5)},anchor=south},
	xlabel={True horizontal length (m)},
	ylabel={Attenuation (dB)},
	] 
	\addplot graphics[xmin=0,xmax=240,ymin=-0.8,ymax=0.8] {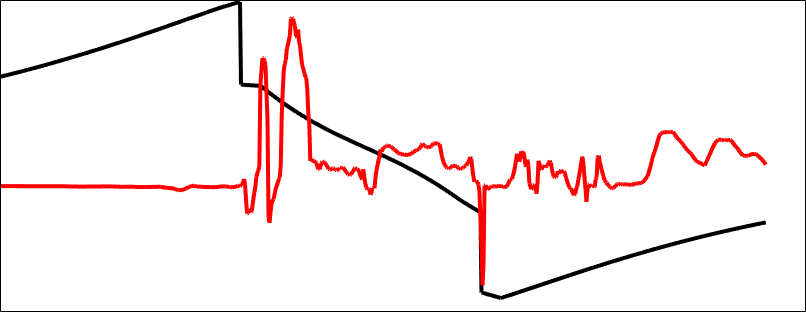};
	\end{axis}
    \node[align=center] (1) at (1.5,1.9)  { {\color{black} Exact}\\ {\color{red} Prediction}};
	\end{tikzpicture} %
			}
			\caption{Model problem 2. Comparison between exact and prediction of attenuation of the measurements.}
			\label{fig_M2_atten}
		\end{figure*}
		\begin{figure*}[ht]
			\centering
			\subcaptionbox{${\cal M}_1$}{%
				\begin{tikzpicture}
	\begin{axis}[
	width=10cm,
	height=3 cm,
		scale only axis,
		enlargelimits=false,
		x label style={at={(axis description cs:0.5,0)},anchor=north},
		y label style={at={(axis description cs:0,0.5)},anchor=south},
		xlabel={True horizontal length (m)},
		ylabel={Phase difference (degree)},
		] 
		\addplot graphics[xmin=0,xmax=240,ymin=-0.2,ymax=0.7] {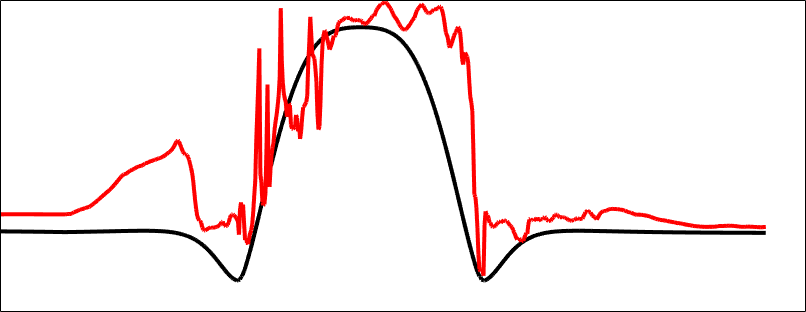};
	\end{axis}
    \node[align=center] (1) at (4.5,1.1)  { {\color{black} Exact}\\ {\color{red} Prediction}};
\end{tikzpicture}    
			}

			\bigskip
			\subcaptionbox{${\cal M}_2$}{%
				\begin{tikzpicture}
	\begin{axis}[
	width=10cm,
	height=3 cm,
		scale only axis,
		enlargelimits=false,
		x label style={at={(axis description cs:0.5,0)},anchor=north},
		y label style={at={(axis description cs:0,0.5)},anchor=south},
		xlabel={True horizontal length (m)},
		ylabel={Phase difference (degree)},
		] 
		\addplot graphics[xmin=0,xmax=240,ymin=-0.15,ymax=0.3] {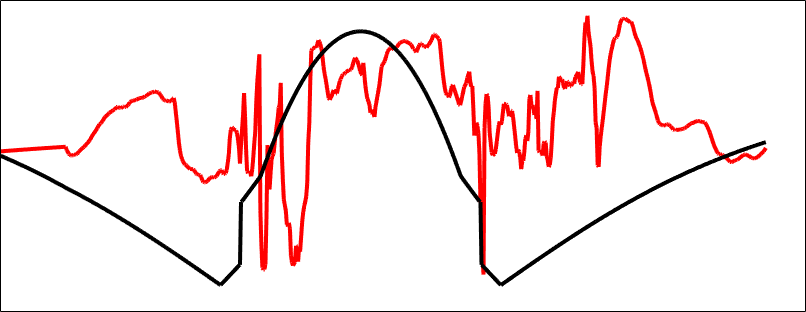};
	\end{axis}
    \node[align=center] (1) at (5.,.75)  { {\color{black} Exact}\\ {\color{red} Prediction}};
\end{tikzpicture}    
			}

			\bigskip
			\subcaptionbox{${\cal M}_3$}{%
				\begin{tikzpicture}
	\begin{axis}[
	width=10cm,
	height=3 cm,
		scale only axis,
		enlargelimits=false,
		x label style={at={(axis description cs:0.5,0)},anchor=north},
		y label style={at={(axis description cs:0,0.5)},anchor=south},
		xlabel={True horizontal length (m)},
		ylabel={Phase difference (degree)},
		] 
		\addplot graphics[xmin=0,xmax=240,ymin=-0.5,ymax=0.6] {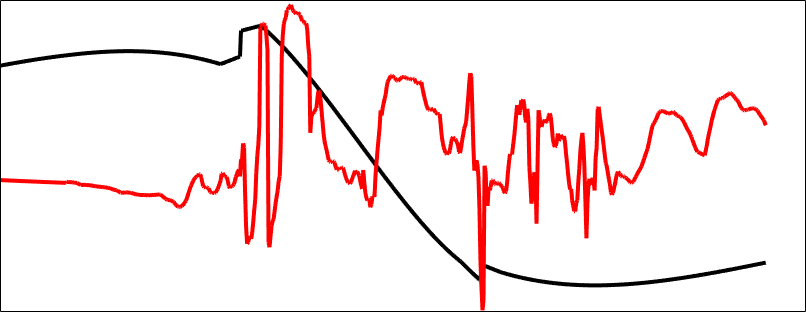};
	\end{axis}
    \node[align=center] (1) at (1.5,1.9)  { {\color{black} Exact}\\ {\color{red} Prediction}};
\end{tikzpicture}    
			}
			\caption{Model problem 2. Comparison between exact and prediction of phase difference of the measurements.}
			\label{fig_M2_phase}
		\end{figure*}
		
		Figure \ref{fig_M2} displays an inversion performed on a three-layer media in which the middle layer is isotropic and also the most resistive one. We consider the other two layers to be anisotropic. As in the previous model problem, results show discrepancies in the anisotropic layers probably because of the lack of a good approximation of anisotropy factor $a$. Figures \ref{fig_M2_atten} and \ref{fig_M2_phase} compare the measurements corresponding to the exact and predicted (inverted) models.
		
		\begin{figure*}[ht]
			\centering
			\subcaptionbox{Actual formation}{%
				\includegraphics[scale=0.3]{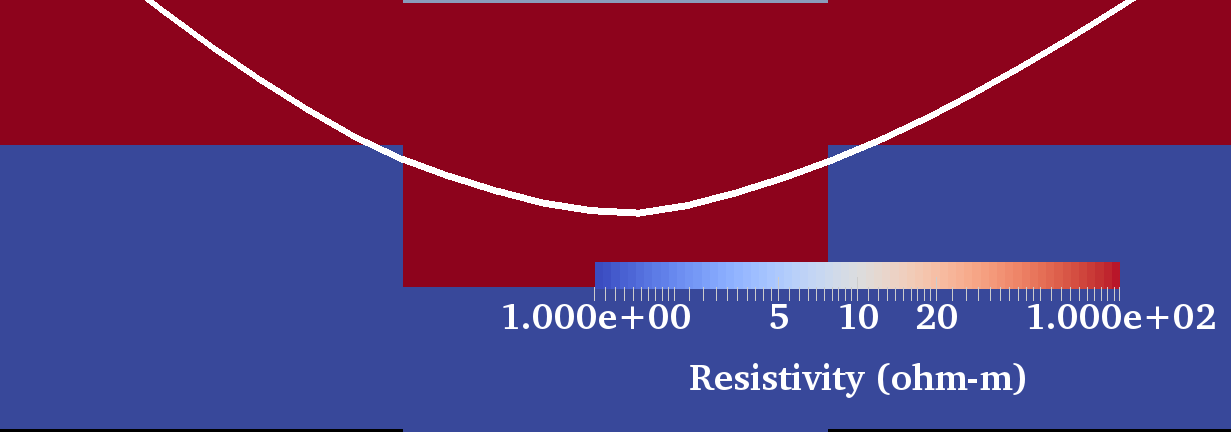} %
			}

			\bigskip
			\subcaptionbox{Predicted (inverted) formation}{%
				\includegraphics[scale=0.3]{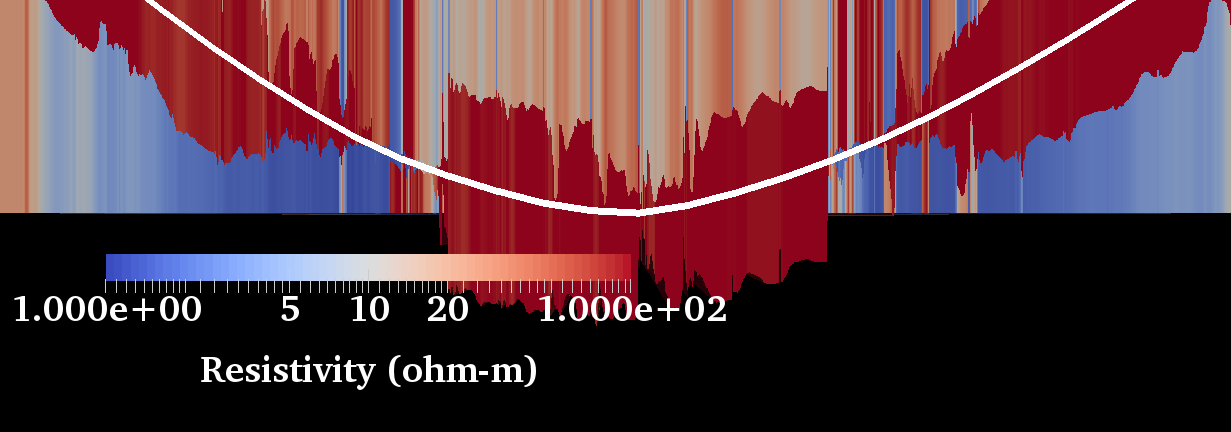}    
			}
			\caption{Model problem 3. Comparison between actual and predicted (inverted) formation.}
			\label{fig_M3}
		\end{figure*}
		\begin{figure*}
			\centering
			\subcaptionbox{${\cal M}_1$}{%
					\begin{tikzpicture}
	\begin{axis}[
	width=10cm,
	height=3 cm,
	scale only axis,
	enlargelimits=false,
	x label style={at={(axis description cs:0.5,0)},anchor=north},
	y label style={at={(axis description cs:0,0.5)},anchor=south},
	xlabel={True horizontal length (m)},
	ylabel={Attenuation (dB)},
	] 
	\addplot graphics[xmin=0,xmax=550,ymin=1.05,ymax=1.6] {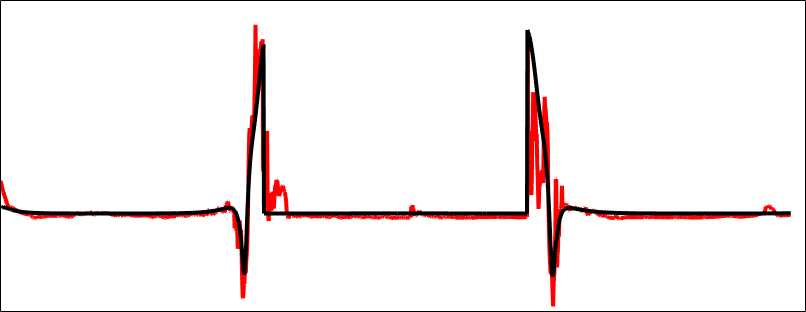};
	\end{axis}
    \node[align=center] (1) at (5,2.2)  { {\color{black} Exact}\\ {\color{red} Prediction}};
	\end{tikzpicture} %
			}

			\bigskip
			\subcaptionbox{${\cal M}_2$}{%
					\begin{tikzpicture}
	\begin{axis}[
	width=10cm,
	height=3 cm,
	scale only axis,
	enlargelimits=false,
	x label style={at={(axis description cs:0.5,0)},anchor=north},
	y label style={at={(axis description cs:0,0.5)},anchor=south},
	xlabel={True horizontal length (m)},
	ylabel={Attenuation (dB)},
	] 
	\addplot graphics[xmin=0,xmax=550,ymin=-7.52,ymax=-7.32] {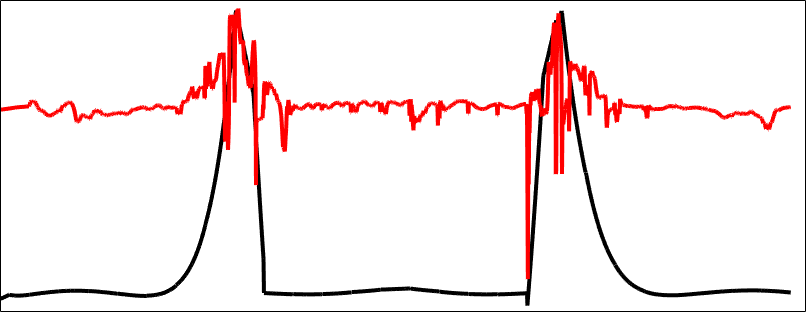};
	\end{axis}
    \node[align=center] (1) at (5,1.)  { {\color{black} Exact}\\ {\color{red} Prediction}};
	\end{tikzpicture} %
			}

			\bigskip
			\subcaptionbox{${\cal M}_3$}{%
					\begin{tikzpicture}
	\begin{axis}[
	width=10cm,
	height=3 cm,
	scale only axis,
	enlargelimits=false,
	x label style={at={(axis description cs:0.5,0)},anchor=north},
	y label style={at={(axis description cs:0,0.5)},anchor=south},
	xlabel={True horizontal length (m)},
	ylabel={Attenuation (dB)},
	] 
	\addplot graphics[xmin=0,xmax=550,ymin=-0.8,ymax=1.2] {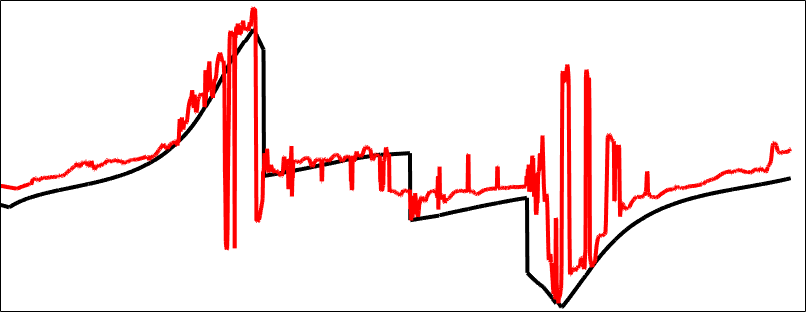};
	\end{axis}
    \node[align=center] (1) at (5,2.2)  { {\color{black} Exact}\\ {\color{red} Prediction}};
	\end{tikzpicture} %
			}
			\caption{Model problem 3. Comparison between exact and prediction of attenuation of the measurements.}
			\label{fig_M3_atten}
		\end{figure*}
		\begin{figure*}[ht]
			\centering
			\subcaptionbox{${\cal M}_1$}{%
				\begin{tikzpicture}
	\begin{axis}[
	width=10cm,
	height=3 cm,
		scale only axis,
		enlargelimits=false,
		x label style={at={(axis description cs:0.5,0)},anchor=north},
		y label style={at={(axis description cs:0,0.5)},anchor=south},
		xlabel={True horizontal length (m)},
		ylabel={Phase difference (degree)},
		] 
		\addplot graphics[xmin=0,xmax=550,ymin=-0.2,ymax=0.7] {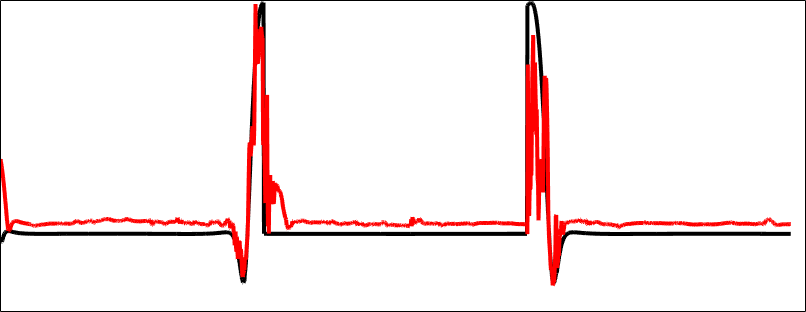};
	\end{axis}
    \node[align=center] (1) at (5,2.2)  { {\color{black} Exact}\\ {\color{red} Prediction}};
\end{tikzpicture}    
			}

			\bigskip
			\subcaptionbox{${\cal M}_2$}{%
				\begin{tikzpicture}
	\begin{axis}[
	width=10cm,
	height=3 cm,
		scale only axis,
		enlargelimits=false,
		x label style={at={(axis description cs:0.5,0)},anchor=north},
		y label style={at={(axis description cs:0,0.5)},anchor=south},
		xlabel={True horizontal length (m)},
		ylabel={Phase difference (degree)},
		] 
		\addplot graphics[xmin=0,xmax=550,ymin=-0.3,ymax=0.7] {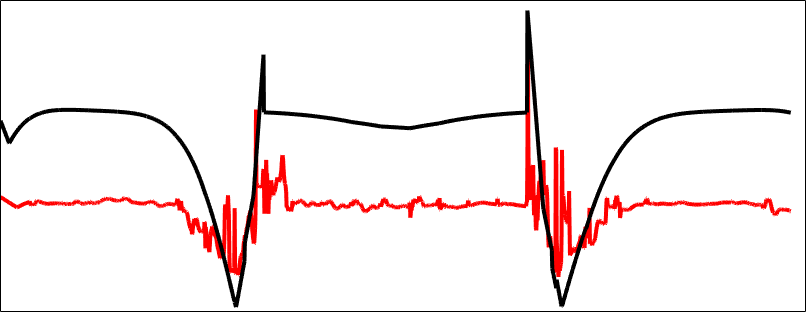};
	\end{axis}
    \node[align=center] (1) at (5,2.4)  { {\color{black} Exact}\\ {\color{red} Prediction}};
\end{tikzpicture}    
			}

			\bigskip
			\subcaptionbox{${\cal M}_3$}{%
				\begin{tikzpicture}
	\begin{axis}[
	width=10cm,
	height=3 cm,
		scale only axis,
		enlargelimits=false,
		x label style={at={(axis description cs:0.5,0)},anchor=north},
		y label style={at={(axis description cs:0,0.5)},anchor=south},
		xlabel={True horizontal length (m)},
		ylabel={Phase difference (degree)},
		] 
		\addplot graphics[xmin=0,xmax=550,ymin=-0.4,ymax=0.5] {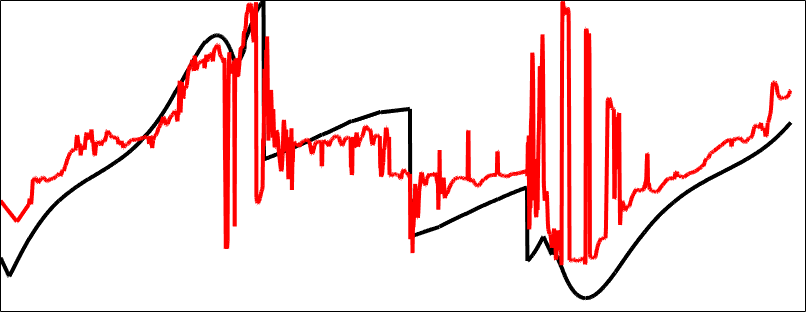};
	\end{axis}
    \node[align=center] (1) at (5,2.45)  { {\color{black} Exact}\\ {\color{red} Prediction}};
\end{tikzpicture}    
			}
			\caption{Model problem 3. Comparison between exact and prediction of phase difference of the measurements.}
			\label{fig_M3_phase}
		\end{figure*}
		Figure \ref{fig_M3} describes the inversion results performed on a synthetic example containing a sequence of 1D layered media. Each 1D model consists of four layers. Inverted results show visible imperfections, and the lack of accuracy for anisotropy factor $a$ causes a poor approximation of the resistivity value in the anisotropic layer.  Predicted $d_u$ and $d_l$ could be employed as a first approximation of the formation surrounding the logging instrument, although a better estimation of $d_u$ and $d_l$ is necessary for a more accurate indicator of the bed boundary positions. Figures \ref{fig_M3_atten} and \ref{fig_M3_phase} compare the measurements corresponding to the exact and predicted (inverted) models. As in previous results, the best approximation is exhibited in the logs corresponding to ${\cal M}_1$. 
		
		\begin{figure*}[ht]
			\centering
			\subcaptionbox{Actual formation}{%
				\includegraphics[scale=0.3]{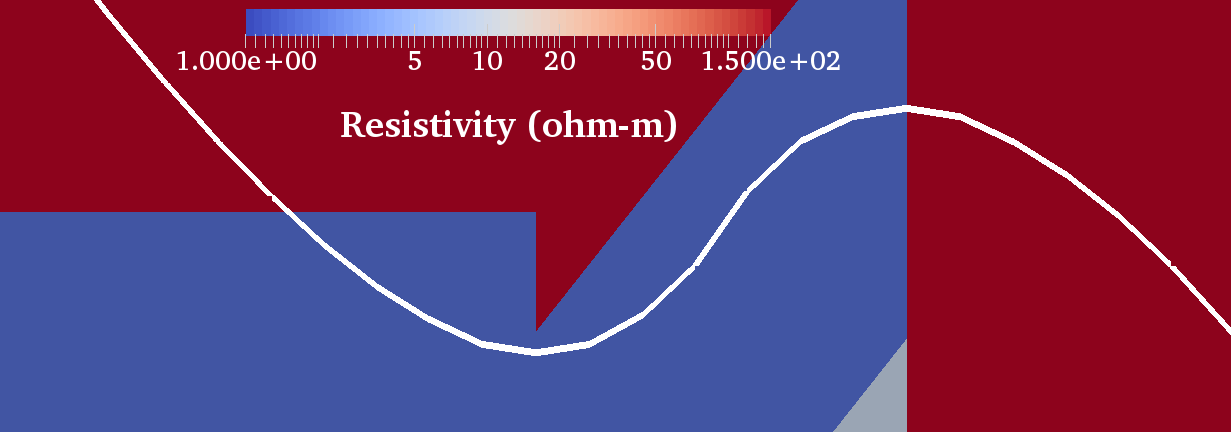} %
			}

			\bigskip
			\subcaptionbox{Predicted (inverted) formation}{%
				\includegraphics[scale=0.3]{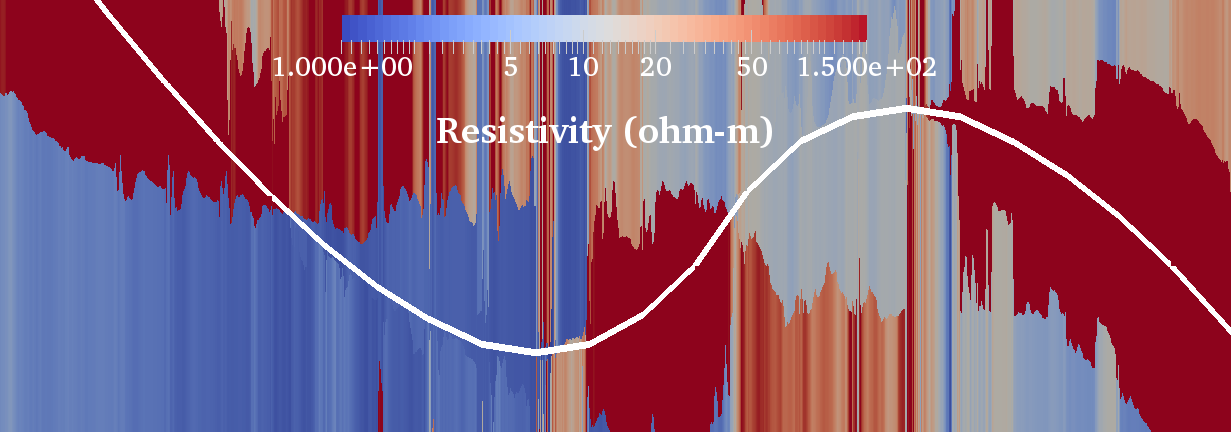}    
			}
			\caption{Model problem 4. Comparison between actual and predicted (inverted) formation.}
			\label{fig_M4}
		\end{figure*} 
		\begin{figure*}[ht]
			\centering
			\subcaptionbox{${\cal M}_1$}{%
					\begin{tikzpicture}
	\begin{axis}[
	width=10cm,
	height=3 cm,
	scale only axis,
	enlargelimits=false,
	x label style={at={(axis description cs:0.5,0)},anchor=north},
	y label style={at={(axis description cs:0,0.5)},anchor=south},
	xlabel={True horizontal length (m)},
	ylabel={Attenuation (dB)},
	] 
	\addplot graphics[xmin=0,xmax=550,ymin=1.,ymax=1.7] {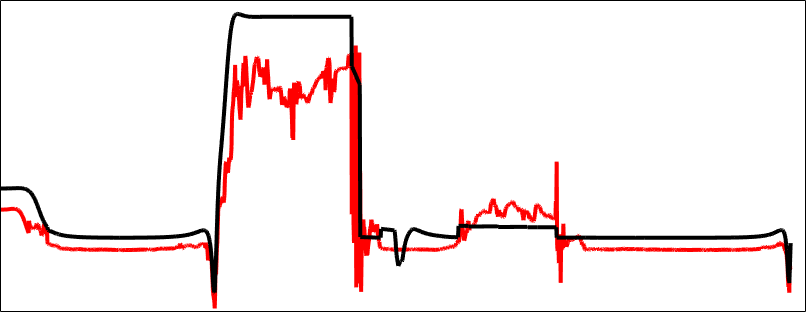};
	\end{axis}
    \node[align=center] (1) at (6.,2)  { {\color{black} Exact}\\ {\color{red} Prediction}};
	\end{tikzpicture} %
			}

			\bigskip
			\subcaptionbox{${\cal M}_2$}{%
					\begin{tikzpicture}
	\begin{axis}[
	width=10cm,
	height=3 cm,
	scale only axis,
	enlargelimits=false,
	x label style={at={(axis description cs:0.5,0)},anchor=north},
	y label style={at={(axis description cs:0,0.5)},anchor=south},
	xlabel={True horizontal length (m)},
	ylabel={Attenuation (dB)},
	] 
	\addplot graphics[xmin=0,xmax=550,ymin=-7.9,ymax=-7.35] {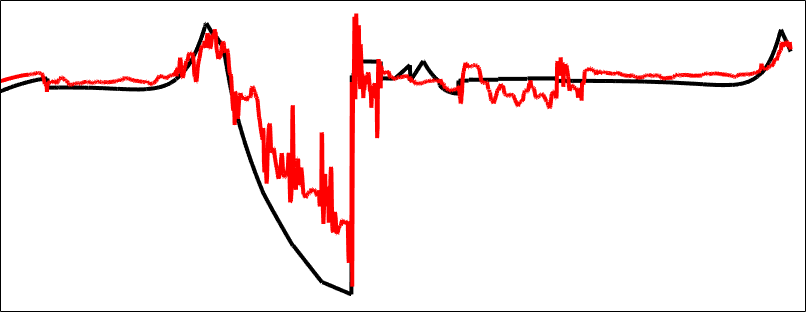};
	\end{axis}
    \node[align=center] (1) at (6,1.3)  { {\color{black} Exact}\\ {\color{red} Prediction}};
	\end{tikzpicture} %
			}

			\bigskip
			\subcaptionbox{${\cal M}_3$}{%
					\begin{tikzpicture}
	\begin{axis}[
	width=10cm,
	height=3 cm,
	scale only axis,
	enlargelimits=false,
	x label style={at={(axis description cs:0.5,0)},anchor=north},
	y label style={at={(axis description cs:0,0.5)},anchor=south},
	xlabel={True horizontal length (m)},
	ylabel={Attenuation (dB)},
	] 
	\addplot graphics[xmin=0,xmax=550,ymin=-0.6,ymax=1.] {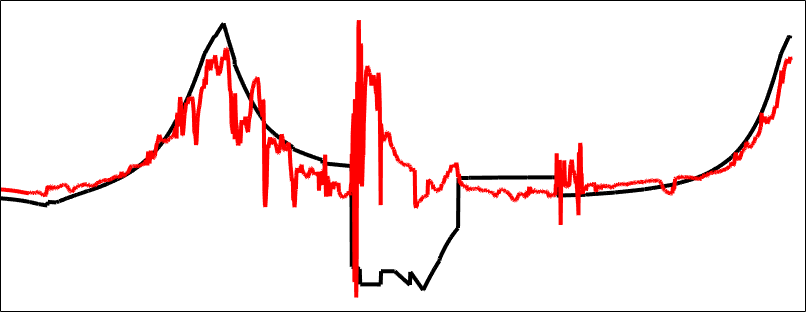};
	\end{axis}
    \node[align=center] (1) at (6.5,2.2)  { {\color{black} Exact}\\ {\color{red} Prediction}};
	\end{tikzpicture} %
			}
			\caption{Model problem 4. Comparison between exact and prediction of attenuation of the measurements.}
			\label{fig_M4_atten}
		\end{figure*}
		\begin{figure*}[ht]
			\centering
			\subcaptionbox{${\cal M}_1$}{%
				\begin{tikzpicture}
	\begin{axis}[
	width=10cm,
	height=3 cm,
		scale only axis,
		enlargelimits=false,
		x label style={at={(axis description cs:0.5,0)},anchor=north},
		y label style={at={(axis description cs:0,0.5)},anchor=south},
		xlabel={True horizontal length (m)},
		ylabel={Phase difference (degree)},
		] 
		\addplot graphics[xmin=0,xmax=550,ymin=-0.2,ymax=0.7] {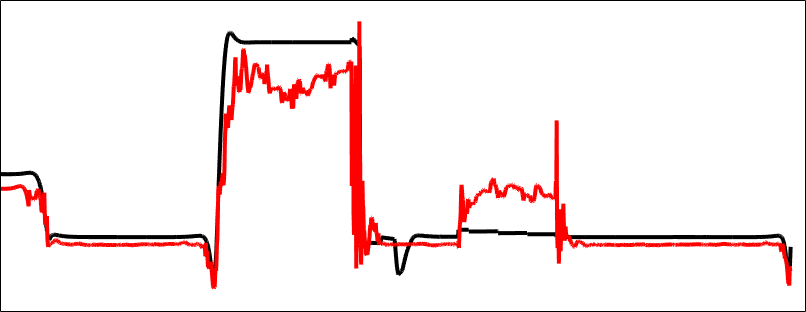};
	\end{axis}
    \node[align=center] (1) at (8.3,2.)  { {\color{black} Exact}\\ {\color{red} Prediction}};
\end{tikzpicture}    
			}

			\bigskip
			\subcaptionbox{${\cal M}_2$}{%
				\begin{tikzpicture}
	\begin{axis}[
	width=10cm,
	height=3 cm,
		scale only axis,
		enlargelimits=false,
		x label style={at={(axis description cs:0.5,0)},anchor=north},
		y label style={at={(axis description cs:0,0.5)},anchor=south},
		xlabel={True horizontal length (m)},
		ylabel={Phase difference (degree)},
		] 
		\addplot graphics[xmin=0,xmax=550,ymin=-0.4,ymax=1.2] {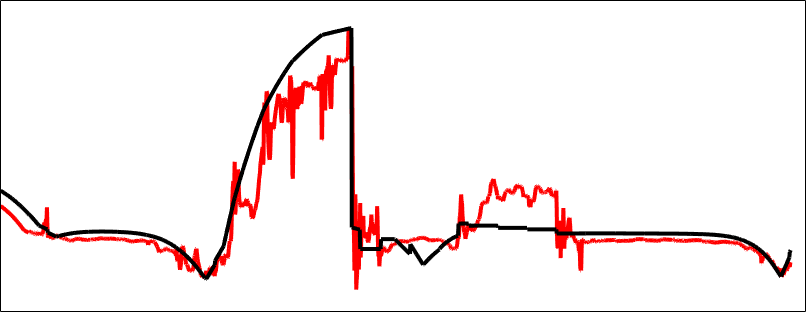};
	\end{axis}
    \node[align=center] (1) at (6.3,2.3)  { {\color{black} Exact}\\ {\color{red} Prediction}};
\end{tikzpicture}    
			}

			\bigskip
			\subcaptionbox{${\cal M}_3$}{%
				\begin{tikzpicture}
	\begin{axis}[
	width=10cm,
	height=3 cm,
		scale only axis,
		enlargelimits=false,
		x label style={at={(axis description cs:0.5,0)},anchor=north},
		y label style={at={(axis description cs:0,0.5)},anchor=south},
		xlabel={True horizontal length (m)},
		ylabel={Phase difference (degree)},
		] 
		\addplot graphics[xmin=0,xmax=550,ymin=-0.4,ymax=0.6] {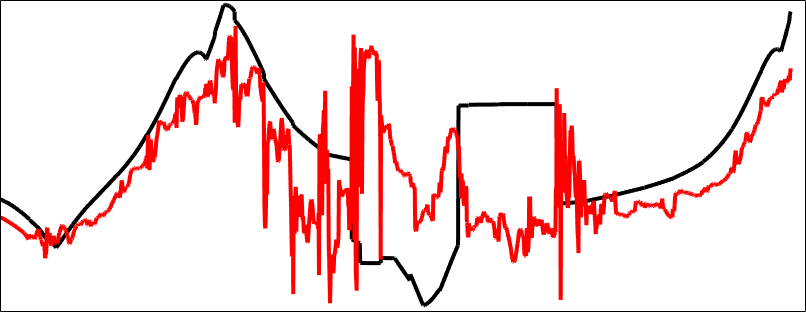};
	\end{axis}
    \node[align=center] (1) at (8.,2.3)  { {\color{black} Exact}\\ {\color{red} Prediction}};
\end{tikzpicture}    
			}
			\caption{Model problem 4. Comparison between exact and prediction of phase difference of the measurements.}
			\label{fig_M4_phase}
		\end{figure*}
		\begin{figure*}[ht]
			\centering
			\subcaptionbox{Actual formation}{%
				\includegraphics[scale=0.3]{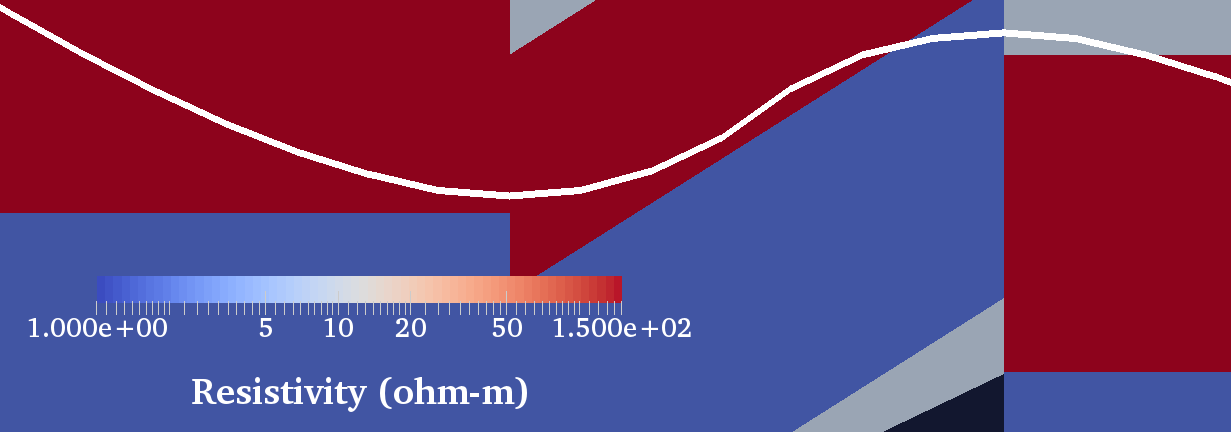} %
			}

			\bigskip
			\subcaptionbox{Predicted (inverted) formation}{%
				\includegraphics[scale=0.3]{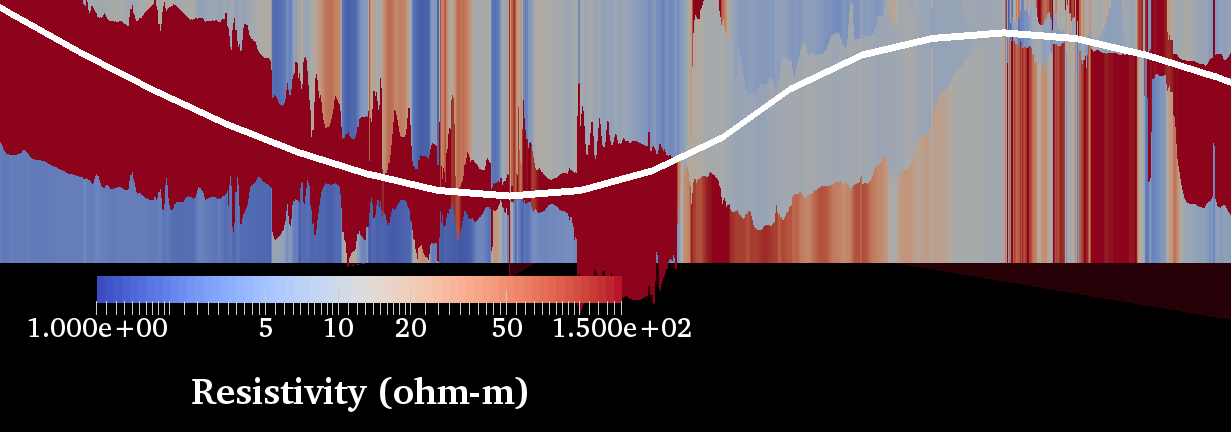}    
			}
			\caption{Model problem 4. Comparison between actual and predicted (inverted) formation.}
			\label{fig_M5}
		\end{figure*} 
		\begin{figure*}[ht]
			\centering
			\subcaptionbox{${\cal M}_1$}{%
					\begin{tikzpicture}
	\begin{axis}[
	width=10cm,
	height=3 cm,
	scale only axis,
	enlargelimits=false,
	x label style={at={(axis description cs:0.5,0)},anchor=north},
	y label style={at={(axis description cs:0,0.5)},anchor=south},
	xlabel={True horizontal length (m)},
	ylabel={Attenuation (dB)},
	] 
	\addplot graphics[xmin=0,xmax=550,ymin=1.1,ymax=1.45] {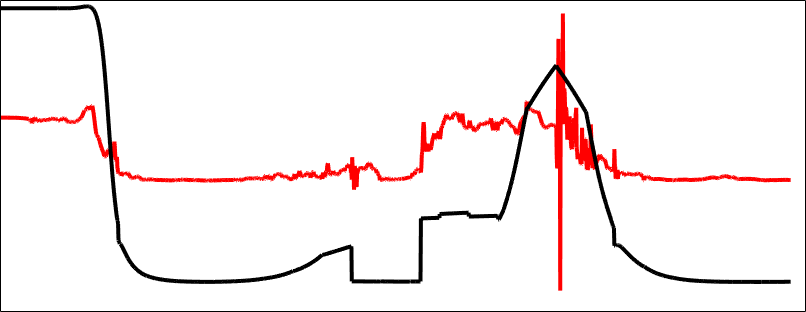};
	\end{axis}
    \node[align=center] (1) at (3.2,2.)  { {\color{black} Exact}\\ {\color{red} Prediction}};
	\end{tikzpicture} %
			}

			\bigskip
			\subcaptionbox{${\cal M}_2$}{%
					\begin{tikzpicture}
	\begin{axis}[
	width=10cm,
	height=3 cm,
	scale only axis,
	enlargelimits=false,
	x label style={at={(axis description cs:0.5,0)},anchor=north},
	y label style={at={(axis description cs:0,0.5)},anchor=south},
	xlabel={True horizontal length (m)},
	ylabel={Attenuation (dB)},
	] 
	\addplot graphics[xmin=0,xmax=550,ymin=-7.56,ymax=-7.4] {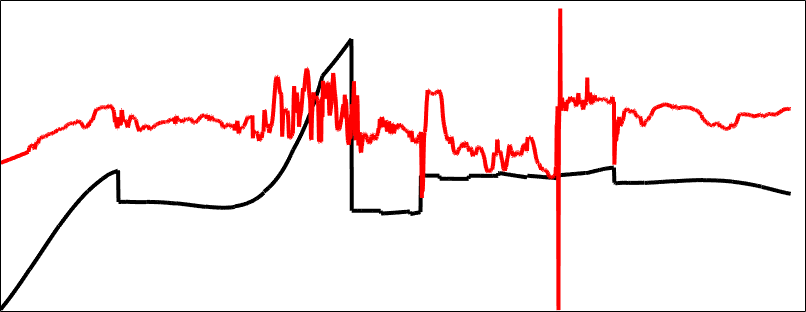};
	\end{axis}
    \node[align=center] (1) at (1.3,2.45)  { {\color{black} Exact}\\ {\color{red} Prediction}};
	\end{tikzpicture} %
			}

			\bigskip
			\subcaptionbox{${\cal M}_3$}{%
					\begin{tikzpicture}
	\begin{axis}[
	width=10cm,
	height=3 cm,
	scale only axis,
	enlargelimits=false,
	x label style={at={(axis description cs:0.5,0)},anchor=north},
	y label style={at={(axis description cs:0,0.5)},anchor=south},
	xlabel={True horizontal length (m)},
	ylabel={Attenuation (dB)},
	] 
	\addplot graphics[xmin=0,xmax=550,ymin=-0.2,ymax=0.7] {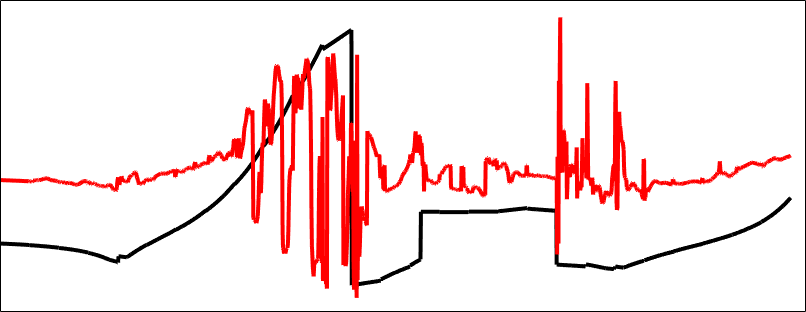};
	\end{axis}
    \node[align=center] (1) at (1.3,2.2)  { {\color{black} Exact}\\ {\color{red} Prediction}};
	\end{tikzpicture} %
			}
			\caption{Model problem 5. Comparison between exact and prediction of attenuation of the measurements.}
			\label{fig_M5_atten}
		\end{figure*}
		\begin{figure*}[ht]
			\centering
			\subcaptionbox{${\cal M}_1$}{%
				\begin{tikzpicture}
	\begin{axis}[
	width=10cm,
	height=3 cm,
		scale only axis,
		enlargelimits=false,
		x label style={at={(axis description cs:0.5,0)},anchor=north},
		y label style={at={(axis description cs:0,0.5)},anchor=south},
		xlabel={True horizontal length (m)},
		ylabel={Phase difference (degree)},
		] 
		\addplot graphics[xmin=0,xmax=550,ymin=-0.2,ymax=0.18] {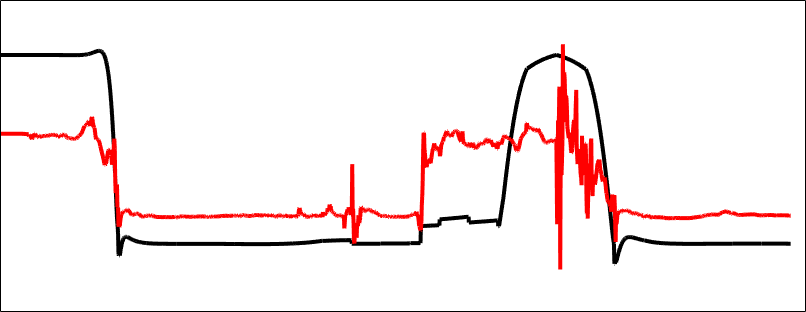};
	\end{axis}
    \node[align=center] (1) at (3,2.2)  { {\color{black} Exact}\\ {\color{red} Prediction}};
\end{tikzpicture}    
			}

			\bigskip
			\subcaptionbox{${\cal M}_2$}{%
				\begin{tikzpicture}
	\begin{axis}[
	width=10cm,
	height=3 cm,
		scale only axis,
		enlargelimits=false,
		x label style={at={(axis description cs:0.5,0)},anchor=north},
		y label style={at={(axis description cs:0,0.5)},anchor=south},
		xlabel={True horizontal length (m)},
		ylabel={Phase difference (degree)},
		] 
		\addplot graphics[xmin=0,xmax=550,ymin=-0.3,ymax=0.5] {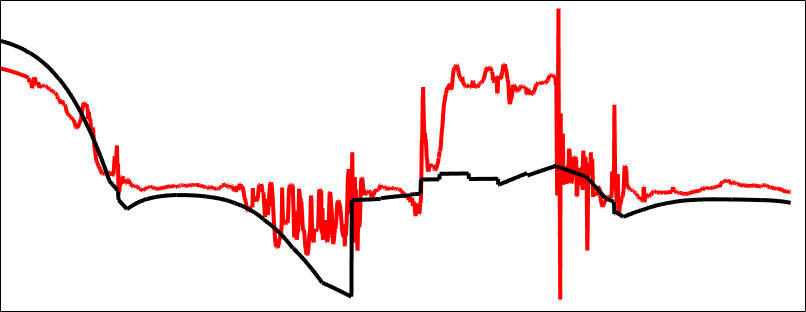};
	\end{axis}
    \node[align=center] (1) at (3,2.2)  { {\color{black} Exact}\\ {\color{red} Prediction}};
\end{tikzpicture}    
			}

			\bigskip
			\subcaptionbox{${\cal M}_3$}{%
				\begin{tikzpicture}
	\begin{axis}[
	width=10cm,
	height=3 cm,
		scale only axis,
		enlargelimits=false,
		x label style={at={(axis description cs:0.5,0)},anchor=north},
		y label style={at={(axis description cs:0,0.5)},anchor=south},
		xlabel={True horizontal length (m)},
		ylabel={Phase difference (degree)},
		] 
		\addplot graphics[xmin=0,xmax=550,ymin=-0.3,ymax=0.5] {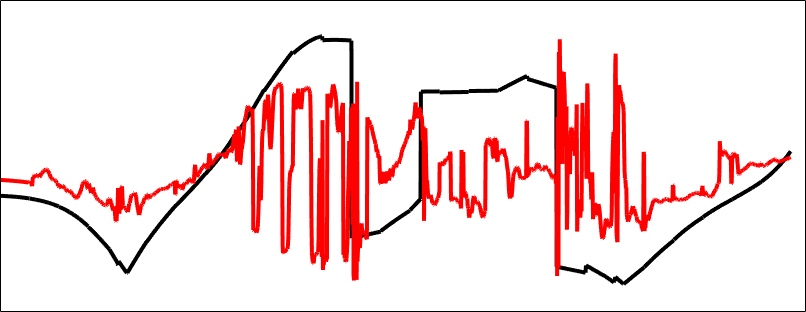};
	\end{axis}
    \node[align=center] (1) at (1.5,2.2)  { {\color{black} Exact}\\ {\color{red} Prediction}};
\end{tikzpicture}    
			}
			\caption{Model problem 5. Comparison between exact and prediction of phase difference of the measurements.}
			\label{fig_M5_phase}
		\end{figure*}
		
		Figures \ref{fig_M4}, \ref{fig_M4_atten}, and \ref{fig_M4_phase} present inversion results for a new synthetic example. A second trajectory is considered to obtain Figures \ref{fig_M5}, \ref{fig_M5_atten}, and \ref{fig_M5_phase}. Again, even if the results present noticeable inaccuracies, DNN results can be used as fast initial approximations that could be refined with other more expensive methods. Notice that DNN inversion results are obtained in a few seconds for over a thousand logging positions and they also provide an uncertainty map.
	}
	\section{Discussion and Conclusion}
	\label{sec:conclusion}
	{\new In this work, we investigated the use of deep learning for the inversion of borehole resistivity measurements. The training stage of a DNN can be a time-consuming stage which can take up to three weeks using GPU to obtain a good approximation. However, we perform the training stage \textit{offline}. Then, the \textit{online} stage (actual inversion) of the method is faster than all other existing conventional inversion methods, which makes it ideal for geosteering purposes. Additionally, using DNNs, we can provide a reliable uncertainty quantification map. Thus, there is an excellent potential in using DNNs for this application.
		
		However, DNNs also present important limitations. First, the inverted results shown in this exploratory work present inaccuracies and further research in the area is still needed. Second, in order to train the system, we require a massive number of data. In the case of a 1.5D model problem, rapid forward solvers exist, which can produce the required data in a reasonable amount of time. However, in the case of 2D and 3D problems, producing such training data set may be extremely time consuming. Moreover, because of the complexity of the problem and the number of variables in the case of 2D and 3D model problems, a much larger data set is required compared to the case of 1.5D problems. Hence, further research is necessary in order to successfully apply DNNs for the inversion of 2D and 3D problems. Third, exploring all possible venues and producing a reliable inversion method using DNNs requires a considerable amount of computational resources and prospective design experimentation. Fourth, the understanding of deep learning algorithms is still limited. In particular, it does not exist a mathematically sound algorithm for the optimal design of the best possible DNN for a given problem. Similarly, it is difficult to recognize a poorly designed DNN. Finally, another limitation of DNNs is that they can only compute a discrete version of the inversion function and when modifying the dimensionality of the measurements, a new DNN should be designed.
		
		The results presented in this work are promising. However, extensive work is still needed in the field to achieve the required accuracy. We envision a large area of research on the topic. As future work, we want to produce more advanced DNNs by designing mechanisms to embed physical constraints associated with our problem into their construction. Furthermore, we want to investigate the use of DNNs for the design of measurement acquisition systems. We can use DNNs for each instrument configuration we design and observe the sensitivity of the desired design to the inversion variables. We shall also investigate the accuracy of the DNNs for noisy data and we will include regularization terms in the cost functional.}
\section{Acknowledgement}
		Mostafa Shahriari and David Pardo have received funding from the European Union's Horizon 2020 research and innovation programme under the Marie Sklodowska-Curie grant agreement No 777778, the Projects of the Spanish Ministry of Economy and Competitiveness with reference MTM2016-76329-R (AEI/FEDER, EU), and MTM2016-81697-ERC/AEI, the BCAM “Severo Ochoa” accreditations of excellence SEV-2013-0323 and SEV-2018-0718, and the Basque Government through the BERC 2014-2017 program, and the Consolidated Research Group Grant IT649-13 on “Mathematical Modeling, Simulation, and Industrial Applications (M2SI)”.
		
		Javier Del Ser thanks the Basque Government for its funding support through the EMAITEK program.
		
		Carlos Torres-Verd\'{i}n was partially funded by The University of Texas at Austin Research Consortium on Formation Evaluation, jointly sponsored by AkerBP, Anadarko, Aramco, Baker-Hughes, BHP Billiton, BP, China, Oilfield Services LTD., Chevron, ConocoPhillips, DEA, ENI, Equinor ASA, Halliburton, Inpex, Lundin-Norway, Nexen, Petrobras, Repsol, Shell, Southwestern, TOTAL, Wintershall and Woodside Petroleum Limited.

	\bibliography{mybibfile}
	\begin{appendices}
		\section{Convolutional Neural Networks}\label{convolutional}
		convolutional neural networks (CNNs) \cite{lecun_gradient-based_1998} are a particular kind of NNs built by replacing fully-connected affine layers $\boldsymbol{\mathcal{N}}$ by convolutional operators $\boldsymbol{\mathcal{C}}$ defined by convolution kernels ${\bf f}$. Hence, Equation \eqref{nn_eq} becomes:
		\begin{equation}\label{cnn_eq}
		\boldsymbol{\cal I}_{\boldsymbol{\theta}}(x) = (\boldsymbol{\mathcal{C}}^{{\bf f}^{(L)}} \circ \ldots \circ \boldsymbol{\mathcal{C}}^{{\bf f}^{(l)}}\circ \ldots \boldsymbol{\mathcal{C}}^{{\bf f}^{(2)}} \circ \boldsymbol{\mathcal{C}}^{{\bf f}^{(1)}})(x),
		\end{equation}
		In a discrete setting, at layer $l$ of Equation \eqref{cnn_eq}, operator $\boldsymbol{\mathcal{C}}^{{\bf f}^{(l)}}$ is determined by the set of convolutional kernels ${\bf f}^{(l)} = \{{\bf f}^{(l)}_s, s=1, \ldots c_{j+1}\}$. 
		Each of these kernels transforms an input tensor ${\bf x}^{(l)}$ of dimension $h_l\times w_l\times c_l$ into an output ${\bf x}_s^{(l+1)}$ of dimension $h_{l}\times w_{l}$. Each kernel is defined by a tensor of dimension $M_l\times N_l\times c_l$ that acts on its inputs through a simple convolution-like operation, followed by a non-linear function like the one in Equation \eqref{activation}:
		\begin{equation}
		\begin{split}
		\displaystyle {\bf x}_s^{(l+1)}(h,w) =\mathbf{s}\Bigg(&\sum_{m=1}^{M_l} \sum_{n=1}^{N_l}  \sum_{c=1}^{c_l} {\bf f}^{(l)}_s(m,n)\\
		& \cdot {\bf x}^{(l)}(h+m, w+n, c)\Bigg).
		\end{split}
		\end{equation}
		Application of all the $c_{l+1}$ convolution kernels of ${\bf f}^{(l)}$ on the input ${\bf x}^{(l)}$ finally results into an output tensor ${\bf x}^{(l+1)}$ of dimension $h_{l}\times w_{l}\times c_{l+1}$.
		Each of these convolutional layers $\boldsymbol{\mathcal{C}}^{{\bf f}^{(l)}}$ is followed by a non-linear point-wise function, and the spatial size of the output from each layer is decreased by a fixed projection operator $\displaystyle\boldsymbol{\mathcal{P}}^{(l)}:\mathbb{R}^{h_l\times w_l} \rightarrow \mathbb{R}^{h_{l+1}\times w_{l+1}}$. Typically, $\boldsymbol{\mathcal{P}}^{(l)}$ is defined as a local averaging operation. Again, eventually the dimensionality of the initial input ${\bf x}$ is transformed into that of an element of the target space $\mathbb{R}^P$.

		\section{Recurrent Neural Networks}\label{recurrent}
\begin{figure*}[ht]
			\centering
			\begin{minipage}[b]{.38\linewidth}
				\centering\includegraphics[width = \textwidth]{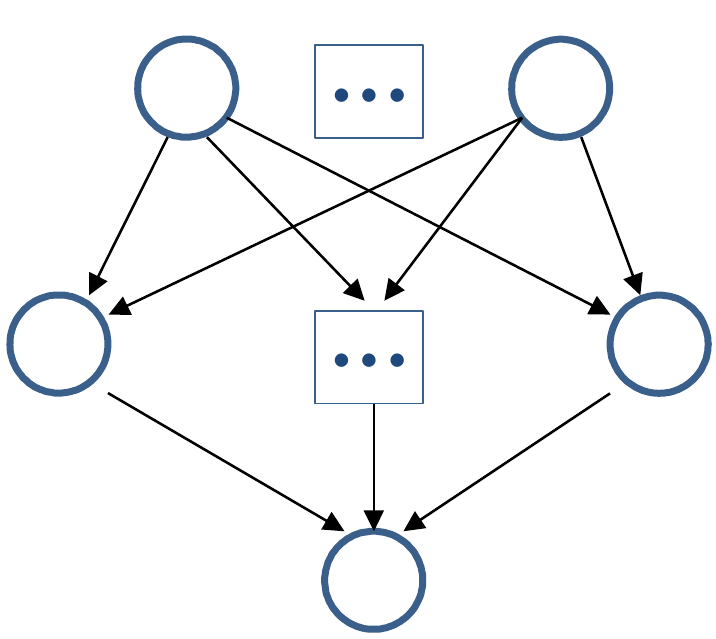}
				\subcaption{Standard NN}\label{fig_nn}
			\end{minipage}%
			\begin{minipage}[b]{.6\linewidth}
				\centering\includegraphics[width=0.8\textwidth]{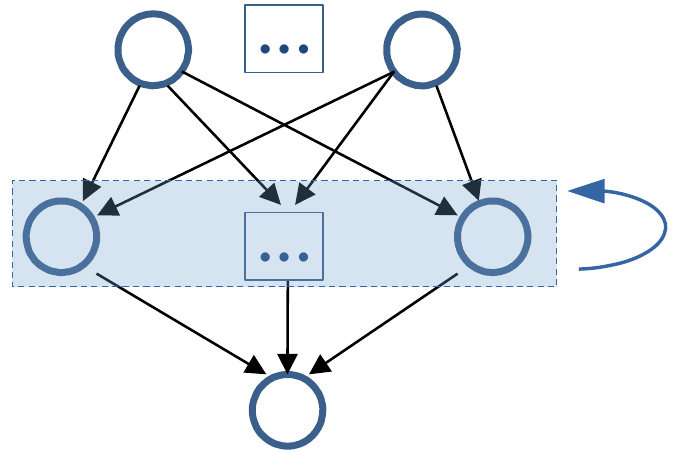}
				\subcaption{Addition of a recurrent connection to (a)}\label{fig_rnn}
			\end{minipage}
			\caption{Comparison between NN and RNN}\label{nn_vs_rnn}
		\end{figure*}

		Let us first consider a simple neural network with an input, an intermediate, and an output layer like the one defined in Section \ref{nn} as a directed graph in which nodes store the result of the operations described in Equation \eqref{nn_eq} and edges store the weights of the network ${\bf W}$, ${\bf b}$, as in Figure \ref{fig_nn}. Computations performed by such a network to obtain an output, given an input ${\bf x}$, are described as:
		\begin{equation}
		\begin{split}
		{\bf z}^{(1)} & =  \mathbf{s}({\bf a}^{(1)}) = \mathbf{s}({\bf W}^{(1)}\cdot {\bf x} + {\bf b}^{(1)}),\\
		\boldsymbol{\cal I}_{\boldsymbol{\theta}}({\bf x})& = \mathbf{s}({\bf W}^{(2)}\cdot {\bf z}^{(1)} + {\bf b}^{(2)}),
		\end{split}
		\end{equation}
		where ${\bf a}^{(1)}$, also known as \textit{activation}, denotes the output of the network at the first layer of this network before passing through the non-linearity $\mathbf{s}$.
		The key difference between regular NN and a recurrent neural network (RNN), as shown in Figure \ref{fig_rnn}, is that the graph defining an NN is acyclical, whereas in an RNN internal cycles are allowed. This introduces a notion of time or sequential dependency into the computations of the network. 
		
		In our case, we interpret a data sample as a temporal sequence of length $T$, ${\bf x}= (x_{1}, x_{2}, ..., x_{T})$, and the goal is to predict an output sequence ${\bf p}$ from ${\bf x}$. In an RNN, a regular NN is trained to predict ${\bf p}=\boldsymbol{\cal I}_{\boldsymbol{\theta}}({\bf x}_{t})$ out of ${\bf x}_{t}$ for $1\leq t \leq T$, but the data is scanned left-to-right, and the previous activation is multiplied by a second set of learnable weights. Hence, the necessary computations within an RNN for a forward pass are specified by the following two equations:
		\begin{equation}\label{rec_eqs}
		\begin{split}
		{\bf a}_{t}&= {\bf W}_{{\bf a}{\bf x}} {\bf x}_{t} + {\bf W}_{{\bf a}{\bf a}}{\bf a}_{t-1} + {\bf b}_{\bf a}\\
		\boldsymbol{\cal I}_{\boldsymbol{\theta}}({\bf x}_{t})&=\mathbf{s}({\bf W}_{{\bf p} {\bf a}} {\bf a}_{t} + {\bf b}_{\bf p}),
		\end{split}
		\end{equation}
		where ${\bf W}_{{\bf a}{\bf x}}$ is a matrix of conventional weights between the input and the inner layer, ${\bf W}_{{\bf a}{\bf a}}$ is a matrix holding recurrent weights between the inner layer at time step $t$ and itself at adjacent time step $t+1$, ${\bf W}_{{\bf a}{\bf x}}$ maps the result of the inner layer computations to the output $\boldsymbol{\cal I}_{\boldsymbol{\theta}}({\bf x}_{t})$, and ${\bf b}_{{\bf a}}, {\bf b}_{\bf p}$ are bias vectors allowing layers within the network to learn an offset. None of the weight matrices depend on the temporal component $t$ and remain fixed, and the transition matrix ${\bf W}_{{\bf a}{\bf a}}$ of the RNN is reset between processing two independent sequences.
		
\begin{figure*}[ht]
			\centering
			\includegraphics[width=\textwidth]{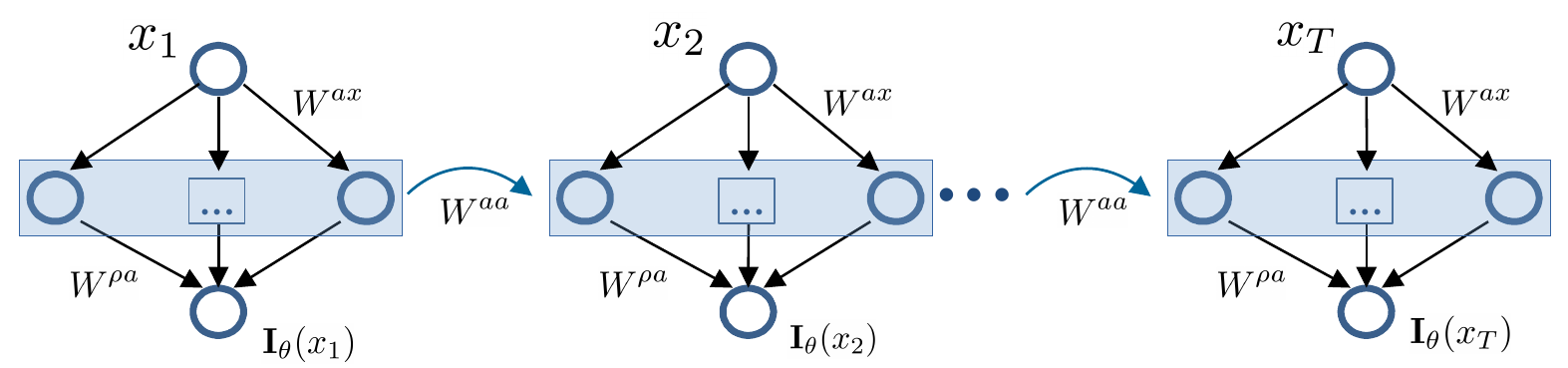}
			\caption{RNN with computations unfolded through time.}
			\label{fig_rnn_unf}
		\end{figure*}

		The temporal nature of the process described in Equation \eqref{rec_eqs} is better illustrated if operations are unfolded, as shown in Figure \ref{fig_rnn_unf}. Following this representation, an RNN can be interpreted not as cyclic, but as a standard network with one layer per time step and shared weights across time steps. It becomes clear that the network can be trained across many time steps using a variant of standard backpropagation algorithm, termed \textit{backpropagation through time} \cite{werbos_backpropagation_1990,hochreiter_gradient_2001}. 
		
		From these first principles, many different flavors of RNNs have been successfully applied over time to temporal data. In this work, we make use of two significant advances in the field of RNNs, namely Long-Short Term Memory RNN (LSTM), and bidirectional recurrent neural network (BRNN). 
		
		LSTM networks \cite{hochreiter_long_1997} are similar to a standard RNN with one inner layer, but a so-called \textit{memory cell} replaces each ordinary node in this layer. Each memory cell contains a node with a self-connected recurrent edge of fixed weight one, ensuring that the gradient can be propagated across many time steps without vanishing or exploding. BRNNs contain two layers, both linked to input and output \cite{schuster_bidirectional_1997}. These two layers are different: the first has a recurrent connection from the past time steps while in the second the direction of recurrent of connections is reversed, performing computations backward along the sequence. More details about both architectures can be found in \cite{lipton_critical_2015}.
		
		\section{Proposed Neural Network Architecture}\label{code}
		The following is a listing of the neural network architecture built in this work in the Keras framework \cite{chollet2015}:
		
		\begin{lstlisting}[language=Python]
		i = Input(shape=input_shape)
		x = LSTM(recurrent_output_size)(i)
		x2 = Reshape((recurrent_output_size,1))(x)
		a = Conv1D(filters=nb_filter, kernel_size=3, activation='relu', kernel_initializer=glorot_normal(),padding='same')(x2)
		d = Conv1D(filters=nb_filter, kernel_size=3, activation='relu',
		kernel_initializer=glorot_normal(),padding='same')(a)
		x = Add()([x2,d])
		x = MaxPooling1D(pool_length)(x)
		a = Conv1D(filters=nb_filter, kernel_size=3,activation='relu', kernel_initializer=glorot_normal(),padding='same')(x)
		d = Conv1D(filters=nb_filter, kernel_size=3, activation='relu', kernel_initializer=glorot_normal(),padding='same')(a)
		x = Add()([x,d])
		x = MaxPooling1D(pool_length)(x)
		a = Conv1D(filters=nb_filter, kernel_size=3, activation='relu', kernel_initializer=glorot_normal(),padding='same')(x)
		d = Conv1D(filters=nb_filter, kernel_size=3, activation='relu',
		kernel_initializer=glorot_normal(),padding='same')(a)
		x = Add()([x,d])
		x = MaxPooling1D(pool_length)(x)
		x= Flatten(input_shape=input_shape)(x)
		y =  Dense(num_outputs, activation='sigmoid', kernel_initializer='glorot_uniform')(x)
		\end{lstlisting}
	\end{appendices}
\end{document}